\documentclass{article}

\usepackage{microtype}
\usepackage{graphicx}
\usepackage{subcaption}
\usepackage{booktabs} % for professional tables
\usepackage[T1]{fontenc}
\usepackage[utf8]{inputenc} %problem with the polnish \k 
\usepackage{hyperref}

\usepackage[accepted]{icml2026}

\usepackage{amsmath}
\usepackage{amssymb}
\usepackage{mathtools}
\usepackage{amsthm}
\usepackage{bm}

\usepackage[capitalize,noabbrev]{cleveref}
\crefname{figure}{Fig.}{Figs.}
\Crefname{figure}{Fig.}{Figs.}
\crefname{table}{Tab.}{Tabs.}
\Crefname{table}{Tab.}{Tabs.}

\theoremstyle{plain}

\theoremstyle{definition}

\theoremstyle{remark}

\usepackage[textsize=tiny]{todonotes}

\usepackage{longtable}
\usepackage{multicol}
\usepackage{makecell}
\usepackage{multirow}
\usepackage{xcolor,colortbl}
\usepackage{array}
\usepackage{soul}
\usepackage{xurl}
\usepackage{enumitem}
\usepackage{float}
\usepackage{wrapfig}

\newcolumntype{G}{>{\columncolor{gray!10}}c}

\usepackage{xspace}
\def\1{\bm{1}}

\def\vb{{\bm{b}}}

\def\vh{{\bm{h}}}

\def\vy{{\bm{y}}}
\def\vz{{\bm{z}}}

\def\mH{{\bm{H}}}

\def\mK{{\bm{K}}}

\def\mQ{{\bm{Q}}}

\def\mV{{\bm{V}}}
\def\mW{{\bm{W}}}

\def\mZ{{\bm{Z}}}

\DeclareMathAlphabet{\mathsfit}{\encodingdefault}{\sfdefault}{m}{sl}
\SetMathAlphabet{\mathsfit}{bold}{\encodingdefault}{\sfdefault}{bx}{n}

\newcommand{\AP}{\texttt{AP}\xspace}
\newcommand{\CLS}{\texttt{CLS}\xspace}
\newcommand{\lastlayer}{\ensuremath{\mathcal{L}_{\text{last}}}\xspace}
\newcommand{\twolayers}{\ensuremath{\mathcal{L}_{\text{mid+last}}}\xspace}
\newcommand{\fourlayers}{\ensuremath{\mathcal{L}_{\text{quarterly}}}\xspace}
\newcommand{\alllayers}{\ensuremath{\mathcal{L}_{\text{all}}}\xspace}

\newcommand{\mm}[1]{}%\textcolor{blue}{[MM: #1]}}
\newcommand{\lc}[1]{}%\textcolor{orange}{[LC: #1]}}
\newcommand{\lm}[1]{}%\textcolor{purple}{[LM: #1]}}
\newcommand{\luca}[1]{}%\textcolor{teal}{[LE: #1]}}

\icmltitlerunning{Attentive Multi-Layer Fusion for Vision Transformers}

\begin{document}

\twocolumn[
\icmltitle{ Attentive Multi-Layer Fusion for Vision Transformers}


\icmlsetsymbol{equal}{*}

\begin{icmlauthorlist}
\icmlauthor{Laure Ciernik}{equal,tub,hfa}
\icmlauthor{Marco Morik}{equal,tub,bifold}
\icmlauthor{Lukas Thede}{tueb,tueai,helm,mcml}
\icmlauthor{Luca Eyring}{helm,mcml,tum}
\icmlauthor{Shinichi Nakajima}{tub,bifold,riken}
\icmlauthor{Zeynep Akata}{helm,mcml,tum}
\icmlauthor{Lukas Muttenthaler}{helm,mcml,tum,aig}
\end{icmlauthorlist}

\icmlaffiliation{tub}{Machine Learning Group, Technische Universit\"at Berlin, Berlin, Germany}
\icmlaffiliation{hfa}{Hector Fellow Academy, Karlsruhe, Germany}
\icmlaffiliation{bifold}{Berlin Institute for the Foundations of Learning and Data (BIFOLD), Berlin, Germany}
\icmlaffiliation{tueb}{University of T\"ubingen, T\"ubingen, Germany}
\icmlaffiliation{tueai}{T\"ubingen AI Center, T\"ubingen, Germany}
\icmlaffiliation{helm}{Helmholtz Munich, Munich, Germany}
\icmlaffiliation{tum}{Technical University Munich, Munich, Germany}
\icmlaffiliation{mcml}{Munich Center for Machine Learning (MCML), Munich, Germany}
\icmlaffiliation{riken}{RIKEN AIP, Tokyo, Japan}
\icmlaffiliation{aig}{Aignostics GmbH, Berlin, Germany}

\icmlcorrespondingauthor{Laure Ciernik}{ciernik@tu-berlin.de}
\icmlcorrespondingauthor{Marco Morik}{m.morik@tu-berlin.de}
\icmlcorrespondingauthor{Lukas Muttenthaler}{lukas.muttenthaler@tu-berlin.de}

% Keywords for PDF metadata
\icmlkeywords{Vision Transformers, Transfer Learning, Probing, Attention, Foundation Models}

\vskip 0.3in
]

% This command creates the footnote with affiliations
\printAffiliationsAndNotice{\icmlEqualContribution}

\begin{abstract}
With the rise of large-scale foundation models, efficiently adapting them to downstream tasks remains a central challenge. Linear probing, which freezes the backbone and trains a lightweight head, is computationally efficient but often restricted to last-layer representations. We show that task-relevant information is distributed across the network hierarchy rather than encoded solely in the last layers. To leverage this distribution of information, we apply an attentive probing mechanism that dynamically fuses representations from all layers of a Vision Transformer. This attentive layer fusion (ALF) learns to identify the most relevant layers for a target task and combines low-level structural cues with high-level semantic abstractions. Across 20 diverse datasets and multiple pretrained foundation models, ALF achieves consistent, substantial gains over standard linear probes.
Attention heatmaps further reveal that tasks different from the pre-training domain benefit most from intermediate representations. Overall, our findings underscore the value of intermediate layers and demonstrate a principled, task-aware approach for unlocking their potential for probing-based adaptation.
\end{abstract}
% Lay summary:
% Modern AI vision models process images through a series of computational stages, called "layers". Early layers detect simple patterns like edges and textures, while deeper layers build up abstract concepts like "fluffy animal with pointy ears". These models are expensive to build, so ideally, we reuse the same model across many different tasks without modifying it. While it has been known that intermediate layers also carry useful information, the standard approach reads off only the final layer's output.

% We propose a lightweight mechanism that learns, for each new task, which layers matter most and how to combine them, while the underlying model itself remains untouched. This automatic, task-aware fusion consistently outperforms approaches that rely on the final layer alone, especially on tasks that differ substantially from the kinds of images the model was originally trained on.

% Our findings provide a principled way to make use of information from all stages inside a model by automatically identifying which layers are most relevant for a given task. This enables more efficient adaptation of large general-purpose models to specialized applications, particularly when computational resources are limited.

%%%%%%%%%%%%%%%%%%%%   Introduction   %%%%%%%%%%%%%%%%%%%%%%%%%%%%%%%%%%%%%%%%%%%%%%%%%%%%%%%%%%%%
\section{Introduction}
\label{sec:introduction}

Foundation models have transformed machine learning across various domains, ranging from language~\citep{devlin-etal-2019-bert,brown2020gpt3} to vision~\citep{pmlr-v139-radford21a, oquab2023dinov2}, by providing powerful pretrained backbones trained on large, general-purpose datasets. How to use these models most effectively for specific downstream tasks remains a central question.
Although \emph{fine-tuning} and its parameter-efficient variants~\citep[e.g., LoRA;][]{hu2022lora,jia2022vpt,chen2022adaptformer} have been proven to yield strong performance, 
% these methods are computationally expensive and require adapting the backbone during training, which changes the underlying model from general-purpose to task-specific.
these methods, whether by updating existing weights or injecting lightweight trainable modules, require gradient computation through the backbone during training, which can entail computational overhead and shifts the model from general-purpose to task-specific behavior. 
A lighter alternative is (linear) \emph{probing} \citep{razavian2014cnnfeaturesofftheshelfastounding, yosinski2014transferable, kornblith2019betterimagenetmodelstransfer}, where the backbone remains unchanged, and a small head is trained on top of it. Probing is attractive in resource-constrained settings, though its accuracy is typically inferior to that of fine-tuning \citep{kornblith2019betterimagenetmodelstransfer}.

The standard linear probing approach operates exclusively on the final-layer representation, which in Vision Transformers ~\citep[ViTs;][]{dosovitskiy2020vit} is typically represented by the \CLS token. This design implicitly assumes that the \CLS token encodes all task-relevant information. However, recent work challenges this assumption: \citet{chen2024context} show that attentive probing over final-layer patch tokens outperforms \CLS-only approaches by facilitating task-dependent spatial information fusion. Similarly, DINOv2~\citep{oquab2023dinov2} demonstrates that concatenating \CLS tokens from several of the last layers can surpass single-layer methods by exploiting some hierarchical information fusion. Together, these results suggest that information crucial for downstream tasks is distributed across layers and tokens rather than exclusively captured by the final \CLS token representation.

%If dependence on the final layer is a limiting factor, 
A potential solution is to fuse the information distributed across the different levels of model layers. ViTs process information across multiple layers: early layers capture low-level visual patterns and structural cues (e.g., edges, textures), whereas later layers encode high-level semantic concepts aligned with the pre-training objective~\citep{raghu2021vision, dorszewski2025colorsclasses}. When downstream tasks differ from the pre-training domain, the final layer likely discards structural or textural information that remains crucial for the target application, yet this information often persists in intermediate layers.

\begin{figure}[t]
    \centering
    \includegraphics[trim={0 0 0 0.95cm}, clip, width=\columnwidth]{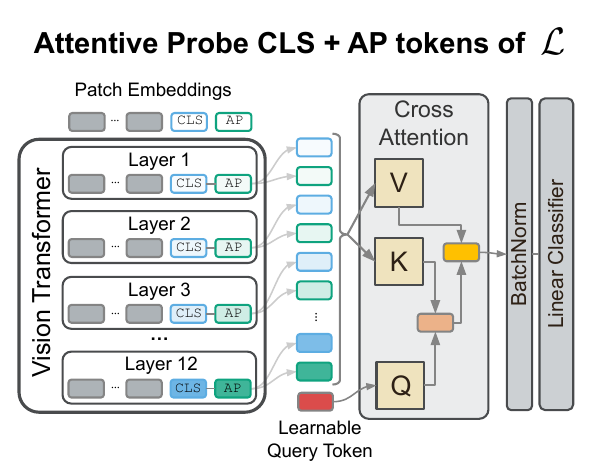}
    \caption{Schematic of our attentive layer fusion (ALF). The method applies cross-attention to \CLS and \AP tokens from multiple transformer layers, automatically discovering which representations contain the most task-relevant features.}
    \label{fig:approach_figure}
    \vspace{-1em}
\end{figure}

Recent work has begun to exploit intermediate representations for transfer learning~\citep{tu2023vqt, wu2024structured,bolya2025perception} and parameter-efficient adaptation~\citep{evci2022head2toe}.
Despite these advances, most approaches rely on simple aggregation strategies, such as concatenation, that produce overly large feature vectors or fail to adapt to task-specific requirements. Furthermore, the relevance of different layers varies substantially across task settings. While some more specialized domains, such as satellite imagery or medical images, benefit from low-level structural cues encoded in early or intermediate layers, others that include a broad set of natural images %(e.g., CIFAR-100) 
require high-level semantic abstraction encoded in the last layers. This variability underscores the need for flexible mechanisms that effectively harness intermediate features in probing settings.

In this work, we demonstrate how to effectively leverage valuable task-relevant information from a model's intermediate layers to substantially improve performance on diverse downstream tasks.
Our evaluation shows that although intermediate layers hold valuable, complementary information, applying standard linear probing to representations from multiple layers is unstable. This indicates a challenge in combining features from a wide range of depths using a simple linear classifier. To solve this, we use an attention-based fusion method that dynamically weights the most informative layers for each task.
Our approach considers both \CLS and average-pooled (\AP) tokens, combining semantic and aggregated spatial information. This method improves performance across various domains and, through the analysis of attention heatmaps, additionally shows how different tasks use the model's hierarchical structure.
Through evaluation across $20$ diverse datasets and vision models from $3$ different families, we find that different tasks adaptively leverage distinct layers of the representational hierarchy.
Hierarchical fusion is particularly effective for datasets outside the pretraining domain and for models that compress information into summary tokens. In contrast, tasks that depend on localized spatial cues benefit from augmenting our approach with complementary spatial aggregation, highlighting the orthogonality between hierarchical and spatial information fusion. \Cref{fig:approach_figure} provides an overview of our proposed attentive layer fusion (ALF).

In summary, our work makes three key contributions:
\begin{itemize}[leftmargin=*,itemsep=0pt,topsep=0pt]
    \item We propose attentive probing using \CLS and \AP tokens from all intermediate layers, achieving consistent gains across 20 datasets, with an average accuracy improvement of 5.54 percentage points over standard linear probing.
    \item We show that intermediate layer fusion provides consistent improvements across small, base, and large models, indicating that the approach generalizes across model scales without diminishing returns.
    \item We find that performance gains are largest for tasks that are different from the pre-training domain. Interpretable attention patterns show that natural image tasks rely more on later layers, whereas structural or specialized datasets benefit more from intermediate representations, particularly the \AP tokens, underscoring the adaptive behavior of our probe.
\end{itemize}

%%%%%%%%%%%%%%%%%%%%   Related Work   %%%%%%%%%%%%%%%%%%%%%%%%%%%%%%%%%%%%%%%%%%%%%%%%%%%%%%%%%%%%
\section{Related Work}
\label{sec:related-works}

ViTs~\citep{dosovitskiy2020vit} pretrained on large-scale datasets, such as CLIP \citep{pmlr-v139-radford21a} and DINOv2~\citep{oquab2023dinov2}, have become fundamental to computer vision. A key challenge is to efficiently use these foundation models for downstream tasks.

\subsection{Probing and Lightweight Adaptation}

Parameter-efficient fine-tuning (PEFT) aims to adapt large neural networks without updating all backbone weights. Popular methods include adapters~\citep{chen2022adaptformer}, visual prompt tuning~\citep{jia2022vpt}, and LoRA~\citep{hu2022lora}.
An even more efficient paradigm is probing, in which the entire pretrained backbone remains frozen and only a lightweight module is trained on top of its features~\citep{DBLP:conf/iclr/AlainB17}. While early work focused on simple linear classifiers, recent studies have introduced more powerful attentive probes~\citep{bardes2024revisiting, el2024scalable}. This idea builds on earlier attention pooling methods, such as the Set Transformer~\citep{lee2019set}, and uses learnable attention modules to aggregate token features from the final layer~\citep{yu2022coca, chen2024context,psomas2025attentionpleaserevisitingattentive}. However, their focus is confined to the final layer's output, implicitly assuming that the final representation is optimal for any given downstream task.

\subsection{The Value of Intermediate Representations}

The principle that hierarchical features are crucial for robust recognition is fundamental to deep learning. In CNNs, representations progress from low-level patterns in early layers to high-level semantics in later ones~\citep{zeiler2014visualizing}. Transferability varies with depth, with earlier layers being more general and later layers more specialized~\citep{yosinski2014transferable}. This led to iconic architectures, such as U-Net~\citep{ronneberger2015unet} and Feature Pyramid Networks \citep{lin2017fpn}, which explicitly fuse features from shallow and deep layers to combine fine-grained details with high-level semantics. This principle extends to ViTs.

Although~\citet{raghu2021vision} showed that their representations are more uniform across layers than in CNNs and representation similarity evolves smoothly over depth~\citep{lange2022}, research confirms that ViT layers still gradually encode more complex semantic concepts~\citep[cf.,][]{ghiasi2022, dorszewski2025colorsclasses}. Recognizing this, architectural extensions such as MViT~\citep{fan2021multiscale}, CrossViT~\citep{chen2021crossvit}, and Swin Transformer~\citep{liu2021swin} explicitly integrate information at different resolutions. More recently, lightweight methods such as Head2Toe~\citep{evci2022head2toe} and Visual Query Tuning~\citep{tu2023vqt} have shown that explicitly exploiting intermediate ViT layers can enhance transfer performance. Similar findings have emerged in language models, where probing has revealed that intermediate layers can even outperform the final layer depending on the task \citep{DBLP:conf/naacl/Liu0BPS19, skean2025layer}. Beyond classification, intermediate representations have also proven valuable for out-of-distribution detection~\citep{imezadelajara2025mysteries} and robotic manipulation~\citep{li2026vatvisionactiontransformer}, further underscoring the generality of this principle across domains and tasks.

Building on these insights, we propose an adaptive attentive probe that learns to dynamically fuse representations from across the entire network hierarchy. Unlike prior work that relies on fixed fusion schemes or manually selected layer subsets, ALF automatically discovers and weights the most task-relevant layers, combining the efficiency of probing with the power of adaptive multi-scale feature fusion.

%%%%%%%%%%%%%%%%%%%%   Method   %%%%%%%%%%%%%%%%%%%%%%%%%%%%%%%%%%%%%%%%%%%%%%%%%%%%%%%%%%%%
\section{Method}
\label{sec:methods}

ViTs learn hierarchical feature representations, with early layers capturing low-level visual patterns and deeper layers encoding high-level semantic concepts \citep{raghu2021vision}. Conventional probing methods, which primarily rely on a model's final layers, may not be optimal for diverse downstream tasks because valuable, task-specific information often resides in intermediate layers. This mismatch necessitates adaptive layer selection to better align the model's feature representations with the specific requirements of the downstream task. We propose an attention-based fusion mechanism that dynamically weights and combines contributions from different transformer layers, allowing the model to automatically find the most relevant features for each downstream task.

\paragraph{Problem Statement.}
Consider a ViT encoder with $L$ attention layers processing an input image $x \in \mathbb{R}^{H \times W \times C}$. For each encoder layer $\ell \in \{1, \dots, L\}$, we extract token embeddings $\mZ^{(\ell)} = [\vz^{(\ell)}_{0}, \vz^{(\ell)}_{1}, \ldots, \vz^{(\ell)}_{P}] \in \mathbb{R}^{(P+1) \times d}$ from the intermediate representation after the second layer normalization, where $\vz^{(\ell)}_{0}$ denotes the \texttt{[CLS]} token representation, $\{\vz^{(\ell)}_{i}\}_{i=1}^{P}$ correspond to $P$ patch-level embeddings, and $d$ is the hidden dimension.

To capture both global and spatial information at each layer $\ell$, we extract two complementary representations, which have been shown to improve performance (see Appx.~\ref{app:ap-token-inclusion}):
\begin{align*}
    \vh^{(\ell)}_{\text{\CLS}} &= \vz^{(\ell)}_{0} ; \quad
    \vh^{(\ell)}_{\text{\AP}} = \frac{1}{P} \sum_{i=1}^{P} \vz^{(\ell)}_{i}.
\end{align*}
The \CLS token provides a learned global summary while average pooling captures spatial feature statistics.
Building on evidence that intermediate layers independently encode valuable task-relevant information (Appx.~\ref{app:intermediate-layer-performance}), we aim to leverage the hierarchical feature evolution across intermediate layers to improve downstream task performance.

Formally, given a subset of layers $\mathcal{L}=\{\ell_1, \ldots, \ell_{|\mathcal{L}|}\} \subseteq \{1, \ldots, L\}$, we stack their representations to form
\begingroup
\setlength{\arraycolsep}{2pt} 
\begin{equation*}
    \mH_\mathcal{L} = \begin{bmatrix} \vh_{\CLS}^{(\ell_1)} & \cdots & \vh_{\CLS}^{(\ell_{|\mathcal{L}|})} & \vh_{\AP}^{(\ell_1)} & \cdots & \vh_{\AP}^{(\ell_{|\mathcal{L}|})} \end{bmatrix}^\top \in \mathbb{R}^{2|\mathcal{L}| \times d}
\end{equation*}
\endgroup

The goal is to learn an attention-based fusion function $f_\theta: \mathbb{R}^{2 |\mathcal{L}| \times d} \rightarrow \mathbb{R}^{d}$ with learnable parameters $\theta$ that produces an optimal task-specific representation.

\subsection{Attention-Based Layer Fusion (ALF)}
To combine representations, we extend the attentive probing paradigm of \citet{chen2024context} from final-layer patch tokens to the full set of intermediate-layer features.

\paragraph{Multi-Head Attention Design.} We employ a multi-head cross-attention mechanism that uses the \CLS and \AP tokens from intermediate transformer layers as input, rather than all final-layer patches as in prior work. ALF attends over the complete set of intermediate representations $H_\mathcal{L}$, enabling task-adaptive selection of optimal abstraction levels.

For each head $m \in \{1, \dots, M\}$, we introduce trainable projection matrices $\mW^{(m)}_{\text{key}}, \mW^{(m)}_{\text{val}}, \mW^{(m)}_{\text{query}}  \in \mathbb{R}^{d \times d_h}$,  with head dimensionality $d_h = 2d / M$.
The shared learnable query matrix $\mQ \in \mathbb{R}^{1 \times d}$ serves as a task-relevance prototype. It adapts during training to prioritize layers containing task-relevant features.
For each head $m$, we compute keys and values from the layer representations $\mH_\mathcal{L}$ and queries from the shared query matrix $\mQ$:
% \begin{align}
%     \mK^{(m)} = \mH_\mathcal{L} \mW^{(m)}_{\text{key}}, \quad \mV^{(m)} = \mH_\mathcal{L} \mW^{(m)}_{\text{val}}, \quad \mQ^{(m)} = \mQ\mW^{(m)}_{\text{query}}
% \end{align}
{\footnotesize
\begin{equation*}
    K^{(m)} = H_{\mathcal{L}}W^{(m)}_{\text{key}}, V^{(m)} = H_{\mathcal{L}}W^{(m)}_{\text{val}}, Q^{(m)} = QW^{(m)}_{\text{query}}
\end{equation*}}
The output of each head is computed as:
{\footnotesize
\begin{equation*}
     \vh^{(m)}_{\text{head}}= \text{dropout}\left(\text{softmax}\left( \frac{\mQ^{(m)} {\mK^{(m)}}^\top}{\sqrt{d_h}} \right)\right) \mV^{(m)},
\end{equation*}
}
where the attention dropout is used as regularization during training. The fused representation is obtained after a linear transformation of the concatenated heads:
\begin{equation*}
\vh_{\text{fused}} = \left[  \vh^{(1)}_{\text{head}}\oplus \cdots \oplus   \vh^{(M)}_{\text{head}} \right]\mW_\text{out} + \vb_{\text{out}}
\end{equation*}

Classification for a downstream task with $K$ classes is performed using a single linear layer with softmax activation $\hat{\vy} = \text{softmax}\!\left( \mW_{\text{clf}} \vh_{\text{fused}} + \vb_{\text{clf}} \right)$.
This design keeps the parameter count independent of the number of layers, with the hidden dimension $d$ being the dominant factor and adds minimal computational overhead, requiring only lightweight attention computations over intermediate representations while keeping the pretrained backbone frozen (Appx. \ref{appx:parameter_count}). 
Although ALF introduces additional attention computations compared to linear probing, the overhead remains moderate relative to full fine-tuning and LoRA (see Appx.~\ref{app:finetuning} for detailed runtime analysis).
The attention computation scales with $\mathcal{O}\left(|\mathcal{L}|^2\right)$ rather than $\mathcal{O}\left(P^2\right)$ for attentive probes on all patches of the last layer. For ImageNet-sized input images, $P \approx 200$, $L \approx 12$, thus, $|\mathcal{L}| \ll P$ yields an order of magnitude reduction in attention complexity.
Rather than manually selecting which layers to include, we find that the use of all intermediate representations $\alllayers = \{1, 2, \ldots, L\}$ performs best, as the learned attention weights automatically determine layer relevance.

\subsection{Linear (Concatenation-Based) Layer Fusion}
To validate the effectiveness of adaptive weighting, we additionally consider the naive approach of combining intermediate representations through concatenation, which we refer to as Linear in our plots. This baseline applies a linear classifier directly to the concatenated features:
\begin{equation*}
     \hat{\vy} = \text{softmax}\!\left( \mW_{\text{clf}}  \left[ \bigoplus _{\ell \in \mathcal{L}} \vh_{\text{CLS}}^{(\ell)} \oplus \vh_{\text{AVG}}^{(\ell)}\right] + \vb_{\text{clf}} \right).
\end{equation*}
This approach leverages intermediate features but lacks ALF's adaptive weighting capability. The importance of each layer's contribution is learned by the single linear classifier but remains fixed for all inputs after training, whereas the attentive probe, in principle, adapts its weighting.

%%%%%%%%%%%%%%%%%%%%   Experiments   %%%%%%%%%%%%%%%%%%%%%%%%%%%%%%%%%%%%%%%%%%%%%%%%%%%%%%%%%%%%
\begin{figure*}[t]
    \centering
    \includegraphics[width=\textwidth]{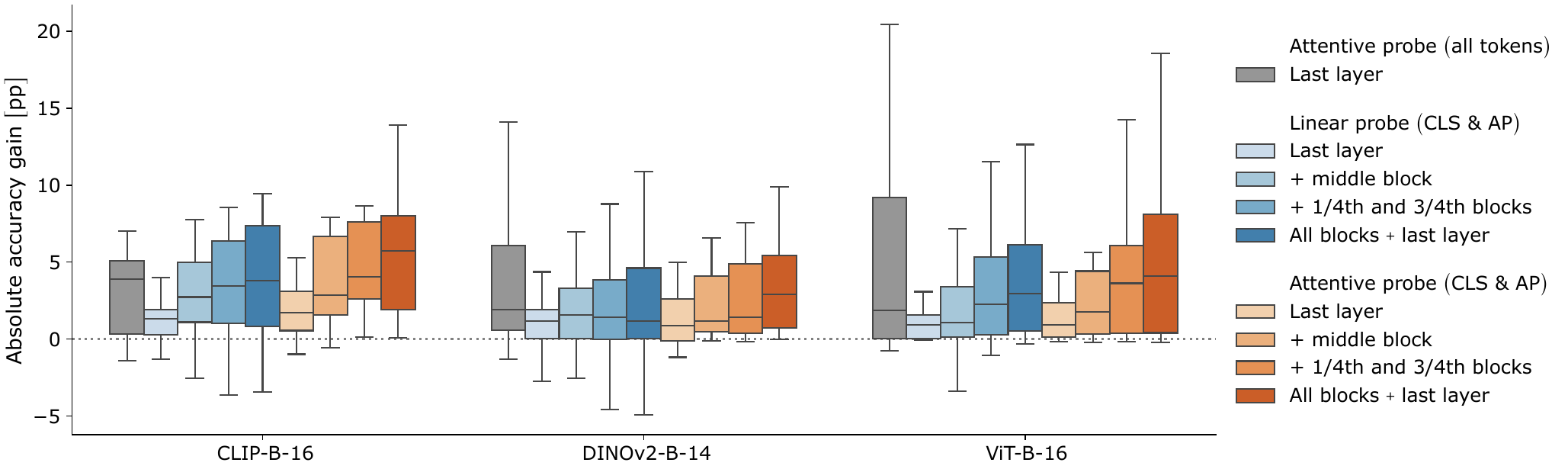}
    \caption{
    Absolute accuracy gain (percentage points) of linear (blue) and attentive probes (orange) when fusing an increasing number of intermediate layer representations (\lastlayer, \twolayers, \fourlayers, and \alllayers), as well as AAT (grey) aggregated across datasets for the three base models. Performance scales with the inclusion of intermediate layers across all models; specifically, ALF (attentive all blocks + last layer) achieves the highest median gain and consistently outperforms the simple linear probe (zero line).
    }
    \label{fig:perf_gain_boxplots_base_models}
\end{figure*}

\section{Experiments} \label{sec:experiments}
To validate attention-based layer fusion, we conduct a large-scale empirical study asking:
(1) Does adaptively fusing intermediate representations outperform strong last-layer baselines?
(2) How do learned fusion strategies vary across architectures, training paradigms, and task domains? We observe consistent improvements from using intermediate layers, with the attention mechanism learning effective layer weights for each downstream task.

\subsection{Experimental Setup}\label{subsec:exp-setup}
We first outline our experimental framework, including the probing methods we compare against and our selection of evaluation models and datasets. For reproducibility, we release our code\footnote{\url{https://github.com/lciernik/attentive-layer-fusion}} and provide further implementation details in Appx.~\ref{app:implementation-details}.

\paragraph{Probing Methods.}
We evaluate probing strategies along three axes: the layer (intermediate vs. last), the tokens (\CLS and/or \AP), and the fusion method (linear vs. attentive). To ensure consistency, we denote probes as [layers] ([tokens], [fusion type]).
We consider two main baselines: Last layer (\CLS, linear), the standard linear probe; and Last layer (all tokens, attentive), an attentive probe \citep{chen2024context} applying multi-head attention over all \lastlayer tokens (AAT).

Our primary approach, ALF, applies attention-based fusion across all layers (\alllayers). For completeness, we also test concatenation and attention over subsets $\mathcal{L}\in{\lastlayer,\twolayers,\fourlayers,\alllayers}$. Here, \twolayers selects the middle and last ViT layers, and \fourlayers selects the last layer from each quarter of a ViT.

\paragraph{Models.}
We evaluate nine pretrained ViTs spanning three families (supervised ViTs, self-supervised DINOv2, and image–text-aligned CLIP), each available in small, base, and large variants, to study the effects of training objective and model capacity. We freeze the backbone, training only the attention-fusion module and classifier on extracted features. See Appx.~\ref{app:model-details} for details. All three families use \CLS tokens in their training objectives, which may compress information into this summary representation. We additionally evaluate Masked Autoencoders~\citep{DBLP:conf/cvpr/HeCXLDG22}, which avoid such compression, in Appx.~\ref{app:mae-results}.

\paragraph{Datasets.} Our evaluation covers a diverse suite of 20 datasets from the clip-benchmark \citep{cherti_2025_15403103} and the Visual Task Adaptation Benchmark (VTAB) \citep{zhai2020largescalestudyrepresentationlearning}. We provide details in Appx. \Cref{tab:datasets_info}.

\paragraph{Training Objective.} To handle class imbalance, we apply a weighted cross-entropy loss~\citep{aurelio_learning_2019}, where the loss for each class is inversely weighted by its number of samples. This weighting scheme balances learning across minority and majority classes. To reduce overfitting in the probing module, we apply standard regularization techniques including attention dropout, weight decay, and light jittering of intermediate representations (see Appx.~\ref{app:implementation-details}).

\paragraph{Evaluation Metric.}
We evaluate model performance using top-1 balanced accuracy on the respective test sets. To enable an intuitive comparison of performances across datasets, we report the absolute accuracy gain (in percentage points [pp]) of each method over the standard linear probe \CLS baseline:
\begin{equation}
    \label{eq:performance_gain}
    \Delta_{\text{Acc}}(\text{method}) = \text{Acc}_{\text{bal}}(\text{method}) - \text{Acc}_{\text{bal}}(\text{CLS}_{\text{linear}}),
\end{equation}
which is positive if the method outperforms the baseline. A single run is reported for each experiment, as preliminary tests indicate small variance across runs with different random seeds (see Appx.~\ref{app:stability-seeds}).

\subsection{The effect of intermediate layers on downstream performance} \label{subsec:do-intermediate-layers-help}
In this section, we aim to assess (1) the impact of adding intermediate layers on the downstream performance compared to using only the final layer, and (2) the performance differences between attentive and linear fusion strategies.
To test this, we evaluate the three base models (CLIP-B-16, DINOv2-B-14, and ViT-B-16) and measure the accuracy gain (Eq.~\ref{eq:performance_gain}) relative to the standard linear probe on the \CLS token as we include more intermediate representations.

\Cref{fig:perf_gain_boxplots_base_models} shows that adding representations boosts performance. Both naive concatenation and ALF significantly benefit from deeper feature pools (p-value $\leq 0.04$, FDR-corrected Wilcoxon, All layers vs Last Layers)
, confirming that intermediate layers encode complementary information absent in the final layer's \CLS and \AP tokens.
However, the two fusion strategies differ substantially in robustness. While concatenation shows positive median gains, it exhibits high variance across tasks, with some datasets experiencing substantial performance degradation.
In contrast, ALF on all layers shows the largest (median) performance gains, consistently outperforming the concatenation fusion strategy (p-value $\leq 0.013$, FDR-corrected Wilcoxon), demonstrating its capacity to adaptively emphasize useful layers while ignoring irrelevant ones.

We compare against the attentive probe on all tokens (AAT) in the last layer, which accesses fine-grained semantic as well as spatial information from all patch tokens (incl. \CLS). AAT proves unstable with high variance and occasional underperformance.
Our method, which attends over summary tokens from all layers, achieves higher median gains with markedly less variance.
We provide detailed results across all datasets and models in~\Cref{app:single_model_complete_results} and comparisons to additional baselines in~\Cref{app:sota_comparison}. 
Finally, we find that our observations are robust across different attentive parameterizations of probes, analyzed in ~\Cref{fig:probe_comparison}.
%We additionally compare against an attentive probe over all tokens in the last layer (AAT), which performs spatial aggregation within a single representational level. In contrast, ALF performs hierarchical aggregation over summary representations across layers. These methods therefore address complementary aspects of the representation space rather than constituting direct alternatives (Appx.~\ref{app:probe_comparison}).

%%% BEFORE ICML rebuttal
% Together, these findings validate two central claims of our work:
% (1) intermediate layers contain valuable information for downstream tasks that is not captured by the final layer alone, and
% (2) an attentive fusion mechanism is crucial to harness this information safely and consistently across diverse downstream tasks.
%%% AFTER ICML rebuttal
Together, these findings validate the central claim of our work: while the value of intermediate layers has been recognized in prior work, we demonstrate that an attentive fusion mechanism is crucial to harness this information safely and consistently across diverse downstream tasks. Naive concatenation proves unstable, revealing that the \emph{manner} of combining layers matters as much as the layers themselves.

\begin{figure*}[h]
    \centering
    \includegraphics[width=\textwidth]{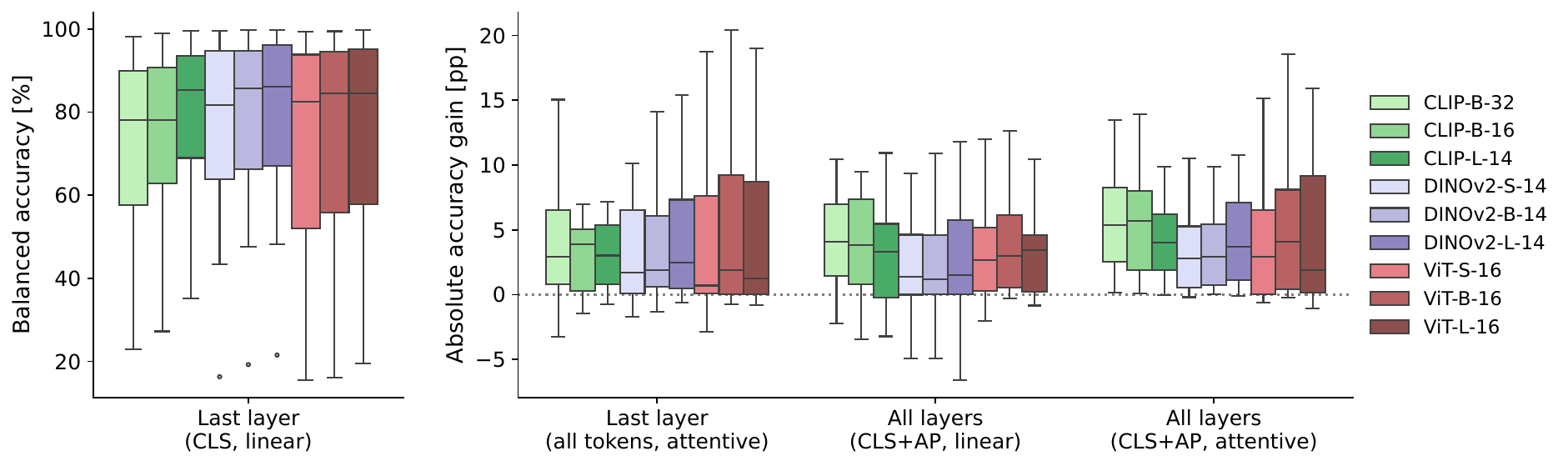}
    \caption{Balanced accuracy distributions of baseline (left panel) and absolute accuracy gains in percentage points (right panel) for different representation fusion methods across model architectures, aggregated over all 20 datasets. The benefits of ALF [All layers (\CLS+\AP, attentive)] persist even for large models, indicating that large models fail to encode all task-relevant information in their final layer's \CLS token.}
    \label{fig:dist_perf_gain_per_model}
\end{figure*}

\subsection{Intermediate layer fusion across model scales}
To assess whether ALF depends on model size, we evaluate small, base, and large variants across three model families.
Consistent with previous results, adding intermediate layers improves performance at all scales (\Cref{fig:dist_perf_gain_per_model}), with attentive probes outperforming concatenation. Detailed per-dataset results for each model are provided in Appx. \Cref{fig:all_models_detailed_perfs}.
However, the magnitude of these gains varies by training objective, yielding distinct scaling behaviors across families.

CLIP models show the most consistent gains, with smaller models (CLIP-B-32/16) benefiting more than the large model, likely because they fail to distill all relevant information into the final layer.

In contrast, DINOv2 models exhibit the opposite trend: performance gains increase with model size, reflecting richer features throughout the network hierarchy.
While the \CLS linear probe already substantially improves from DINOv2-S-14 to DINOv2-L-14, attentive fusion adds a further mean gain of 6.04 [pp] for DINOv2-L-14. 
AAT yields a slightly higher average gain (6.23 [pp]), but suffers from greater instability. % and lower median gain. 
This likely stems from AAT's reliance on hundreds of patch tokens (257 for DINOv2-S/B/L-14), which amplifies task-specific noise. In contrast, ALF aggregates only 24/48 summary tokens, producing more consistent improvements across tasks.
%AAT yields a slightly higher average gain (6.23 [pp]) in this setting, but suffers from greater instability, reflecting the higher variance inherent to spatial token-level aggregation compared to hierarchical fusion over compact summary tokens.

Finally, for supervised ViT models, gains peak at the base architecture. The large variant benefits less proportionally, which could be due to overfitting from the higher hidden dimension or to diminishing returns in representational richness as backbone capacity grows. In either case, while relative improvements shrink, absolute performance still increases when attending across layers.

In summary, attentive fusion consistently improves performance across model sizes. Contrary to the intuition that smaller models would benefit more due to their weaker base performance, we find that larger models obtain equally substantial gains. Highlighting ALF's ability to scale with model capacity, it complements rather than replaces the final-layer representation.
At the same time, the variability across datasets suggests that the benefits are task dependent, which we elaborate on in the following section.

\begin{figure*}[t]
    \centering
    \includegraphics[width=0.95\textwidth]{figures/which_layers_have_been_used_attention_heatmaps/AttentionWeightsBaseModels.png}
    \caption{Attention weights across layers and datasets for base models, averaged over heads and samples, are distributed across multiple layers, demonstrating their relevance for downstream tasks.}
    \label{fig:attn_heatmap_base_models}
\end{figure*}

\subsection{Task dependent benefits of intermediate layer fusion}
\label{sec:exp:task_dependent}
\Cref{tab:perf_gain_per_dataset} summarizes the effect of different probing strategies across the 20 benchmark datasets, averaged over the nine models considered in this work, where ALF achieved the strongest gains (yielding the highest mean rank).
The exact magnitude of the improvements from intermediate layers is somewhat task-dependent. On natural multi-domain datasets (CIFAR-10, STL-10), the baseline accuracy is near saturation, and fusion therefore yields relatively small but still significant gains. Fine-grained natural-image tasks (Stanford Cars, FGVC Aircraft, GTSRB, SVHN) benefit most from attentive probing, with gains of 6–30 [pp]. These datasets require subtle distinctions between visually similar categories or precise spatial reasoning, which the final \CLS token, optimized for global summarization, tends to suppress.
%While AAT surpasses ALF on these particular tasks by leveraging fine-grained spatial cues from patch embeddings, ALF remains the second-best in almost all cases and, importantly, provides the best average performance and best average rank across all 20 datasets (\Cref{tab:perf_gain_per_dataset}). 
While AAT surpasses ALF on these particular tasks by leveraging fine-grained spatial cues from patch embeddings, ALF remains the second-best in almost all cases (\Cref{tab:perf_gain_per_dataset}), reflecting its focus on hierarchical abstraction rather than spatial precision.
Despite potential trade-offs in spatial granularity due to patch aggregation, ALF remains more stable than AAT and consistently beats linear probing, validating the robustness of its multi-layer approach.

\begin{table}[h]
    \centering
    % \caption{Absolute performance gains (\%) relative to baseline linear probe on CLS token from the final layer across benchmark datasets. Values represent mean ± standard deviation computed across all models using both CLS and average pool tokens for each method. \textbf{Bold} indicates the largest performance gain per dataset, and \underline{underlined} indicates the second-largest. Additionally, the baseline mean balanced accuracy across models is shown for each dataset.Dataset categories: Natural (MD) = MD natural images, Natural (SD) = single-domain natural images, Specialized = domain-specific imagery, Structured = datasets with inherent structural patterns.}
    \caption{Absolute performance gains (pp) of probing methods relative to baseline CLS token linear probe on the final layer. Results show mean ± standard deviation across all 9 models. \textbf{Bold} indicates the largest performance gain per dataset, and \underline{underlined} indicates the second-largest. Baseline balanced accuracy reported for reference. Dataset categories top to bottom: Natural multi-domain; Natural single domain; Specialized; Structured.}
    \resizebox{\linewidth}{!}{
    \begin{tabular}{l|c|ccc@{}}
\toprule
 Dataset & \begin{tabular}[t]{@{}c@{}}Baseline \\ Bal. accuracy\\(CLS, linear)\end{tabular} & \begin{tabular}[t]{@{}c@{}}Last layer\\(all tokens,\\ attentive)\end{tabular} & \begin{tabular}[t]{@{}c@{}}All layers\\(CLS+AP,\\linear)\end{tabular} & \begin{tabular}[t]{@{}c@{}}All layers\\(CLS+AP, \\attentive)\end{tabular} \\
\midrule
\midrule
STL-10         & 99.29 ± 0.51 & 0.01 ± 0.16  & \underline{0.03 ± 0.10} &\textbf{0.04 ± 0.17} \\
CIFAR-10       & 96.91 ± 1.93 & 0.42 ± 0.58  & \underline{0.61 ± 0.71} & \textbf{0.77 ± 0.79} \\
Caltech-101    & 95.57 ± 1.40 & 0.23 ± 0.52  & \underline{0.36 ± 0.63} & \textbf{0.88 ± 0.77} \\
PASCAL VOC07   & 87.82 ± 2.31 & -0.22 ± 1.24 & \textbf{1.46 ± 0.99}    & \underline{1.24 ± 0.89} \\
ImageNet-1k    & 81.40 ± 4.49 & 0.85 ± 1.43  & \underline{0.99 ± 1.75} & \textbf{1.24 ± 1.62} \\
CIFAR-100      & 85.45 ± 5.71 & 1.73 ± 1.33  & \underline{2.76 ± 2.48} & \textbf{3.33 ± 2.75} \\
Country-211    & 21.48 ± 6.35 & -0.83 ± 1.66 & \underline{3.26 ± 1.05} & \textbf{4.96 ± 1.37} \\
\midrule
Pets           & \underline{93.98 ± 2.36} & -0.23 ± 0.83            & -2.01 ± 1.04 & \textbf{0.29 ± 0.76} \\
Flowers        & 98.03 ± 2.60             & \underline{0.41 ± 0.93} & -0.25 ± 0.57 & \textbf{0.46 ± 0.97} \\
Stanford Cars  & 77.81 ± 10.65            & \textbf{8.97 ± 5.22}    & -0.86 ± 3.76 & \underline{6.35 ± 3.71} \\
FGVC Aircraft  & 55.69 ± 12.18            & \textbf{9.27 ± 4.37}    & -1.62 ± 5.01 & \underline{6.43 ± 3.25} \\
GTSRB          & 71.51 ± 7.46             & \textbf{18.02 ± 6.37}   & 8.76 ± 4.20  & \underline{13.47 ± 4.92} \\
SVHN           & 56.06 ± 5.91             & \textbf{30.31 ± 5.08}   & 24.40 ± 4.41 & \underline{27.25 ± 4.24} \\
\midrule
PCAM           & 82.04 ± 2.15 & \underline{5.03 ± 1.47} & \textbf{5.32 ± 1.62}    & 2.85 ± 2.53 \\
EuroSAT        & 93.89 ± 2.52 & 3.38 ± 2.18             & \underline{4.08 ± 2.48} & \textbf{4.37 ± 2.41} \\
RESISC45       & 90.45 ± 1.69 & 4.07 ± 1.05             & \underline{4.53 ± 0.99} & \textbf{5.23 ± 1.10} \\
Diab. Retinop. & 45.80 ± 2.46 & 1.94 ± 1.90             & \underline{5.92 ± 2.03} & \textbf{6.86 ± 2.00} \\
\midrule
DTD            & 75.99 ± 3.47 & 1.41 ± 2.19             & \underline{4.04 ± 2.19} & \textbf{4.05 ± 1.92} \\
FER2013        & 59.08 ± 4.61 & \underline{7.74 ± 2.15} & 6.25 ± 1.19             & \textbf{10.05 ± 1.76} \\
Dmlab          & 44.91 ± 3.49 & \textbf{13.69 ± 2.77}   & 7.92 ± 1.95             & \underline{10.68 ± 2.78} \\
\midrule
\midrule
  Mean rank & 3.48 & \underline{2.29} & 2.43 & \textbf{1.33} \\ 
\bottomrule
\end{tabular}
    }
    \label{tab:perf_gain_per_dataset}
    \vspace{-8pt}
\end{table}

Domain-specialized (satellite or medical imagery) and structured datasets (textures, facial expressions, synthetic environments) benefit substantially from the inclusion of intermediate layers, reflecting the transferability of mid-level features to novel domains and their encoding of compositional patterns.
A notable exception is DMLab, where patch-level aggregation performs better, because fine spatial detail is critical for this task.

Beyond mean performance gains, stability matters. 
%Attending to all last-layer tokens can excel on certain fine-grained tasks but is brittle, sometimes degrading performance when the \CLS token already suffices (e.g., Pets). In contrast, our attention over summary tokens from all layers consistently delivers performance gains across all datasets. 
Attending to all last-layer tokens can excel on certain fine-grained tasks but is brittle, as spatial aggregation amplifies token-level noise when fine localization is unnecessary. In contrast, hierarchical aggregation over summary tokens provides a more stable inductive bias across tasks.
Only for PCAM and PASCAL VOC 2007 does the linear combination of intermediate layers outperform attentive weighting, likely due to overfitting, as discussed in Appx.~\ref{app:overfitting}.

Taken together, these results demonstrate that the usefulness of intermediate features varies by task. The benefits are greatest for datasets outside the pretraining domain, where the \CLS token alone often proves to be insufficient.
While outliers such as PCAM reveal the risk of overfitting, adaptive fusion remains the most reliable strategy for exploiting task-specific signals from intermediate layers.

\begin{figure*}[h]
    \centering
    \includegraphics[width=\linewidth]{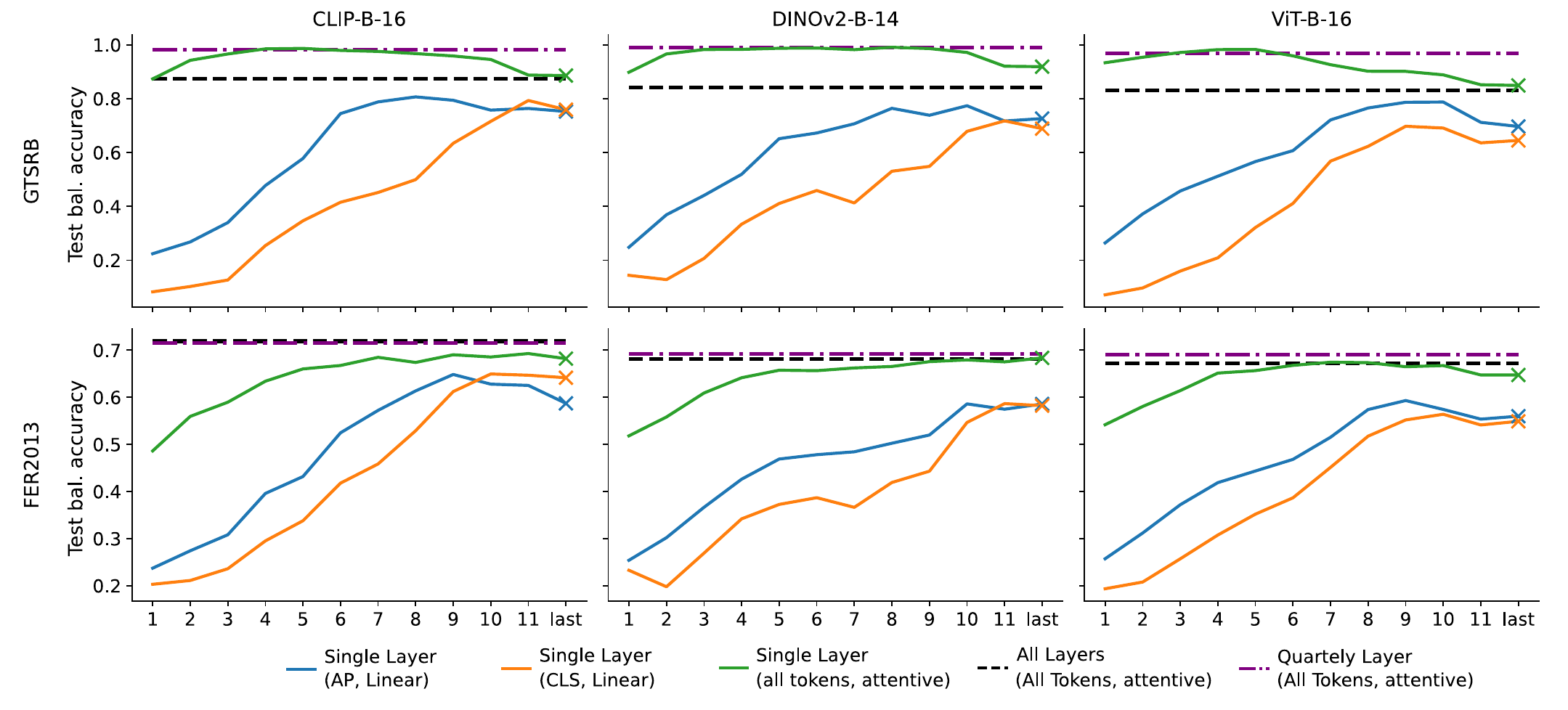}
    \caption{Probing performance on GTSRB (upper row) and FER2013 (lower row) across intermediate layers. Linear probes (single-layer \AP or \CLS) are compared against attentive strategies across CLIP, DINOv2, and ViT-B/16. The black dashed line denotes ALF (hierarchical \CLS+\AP aggregation), while the purple dash-dotted line represents the hybrid approach aggregating all tokens across quarterly layers. Results highlight that while ALF ensures stability, combining spatial and hierarchical fusion yields peak performance.}
    % \vspace{-8pt} 
    \label{fig:intermediate_orthogonal}
\end{figure*}

\subsection{Analyzing adaptive layer selection}
To understand how ALF adapts to different downstream tasks, we analyze the attention weights of intermediate layers. These weights reveal which layer's representations are most crucial for a given dataset. By aggregating over the attention heads and data samples, the heatmaps indicate how much each layer contributes to the fused representation (\Cref{fig:attn_heatmap_base_models} for base and Appx. \Cref{fig:appx:attn_smalllarge} for small/large model sizes).

Early layers' \CLS tokens receive little attention, which is expected since the global summary only becomes semantically rich in later layers. In contrast, average-pooled representations are used across a much wider range of layers. This confirms our hypothesis that spatial averaging preserves valuable textural and structural information throughout the network, complementing the highly processed \CLS tokens. Even when restricting ALF to roughly 50\% of tokens (Appx.~\ref{app:layerpruning}), performance remains close to full inference, indicating that ALF learns the selection of relevant layers without manual choice.

Notably, the attention distribution varies by dataset. For tasks similar to pretraining, like CIFAR or Pets, attention is high on the last layers' \CLS and \AP tokens, as these abstract features are directly useful. In contrast, for tasks that differ from pretraining, such as EuroSAT and FER2013, attention shifts to intermediate layers and their \AP tokens, consistent with the largest performance gains observed on these datasets.
As shown in Appx.~\ref{app:intermediate-layer-alone} and ~\ref{app:intermediate-layer-performance}, intermediate layers alone can achieve performance comparable to that of the last layer, despite having dissimilar representations. This suggests that these layers provide potential complementary, non-redundant information across layers. Overall, the heatmap confirms that adaptive fusion effectively leverages these lower-level features that might otherwise be lost in the last layers.

% \begin{figure}[t]
%     \centering
%     \includegraphics[width=\linewidth]{figures/orthogonality/cka_global_vs_test_bal_acc_intermediate_layers_with_aatilf_only_gtsrb.pdf}
%     \caption{Probing performance on GTSRB across intermediate layers. Linear probes (single-layer \AP or \CLS) are compared against attentive strategies across CLIP, DINOv2, and ViT-B/16. The black dashed line denotes ALF (hierarchical \CLS+\AP aggregation), while the purple dash-dotted line represents the hybrid approach aggregating all tokens across quarterly layers. Results highlight that while ALF ensures stability, combining spatial and hierarchical fusion yields peak performance.}
%     \vspace{-8pt} 
%     \label{fig:intermediate_orthogonal}
% \end{figure}

\subsection{On the orthogonality of spatial and hierarchical information aggregation}
\label{subsec:orthogonality}
For most tasks, our hierarchical fusion using summary tokens (ALF) is sufficient as the average pooling provides some spatial statistics to complement \CLS semantics (\Cref{fig:ap-token-inclusion}). However, for tasks requiring fine-grained spatial precision (e.g., GTSRB), AAT offers additional benefits, as evidenced by the performance gains noted in \Cref{tab:perf_gain_per_dataset}. Importantly, these two mechanisms capture different and complementary aspects of the representation space. When we combine spatial aggregation (across patches) with hierarchical aggregation (across quarterly layers), performance improves beyond using either alone (\Cref{fig:intermediate_orthogonal}, details in \Cref{app:intermediate-layer-performance}). 
This indicates that hierarchical fusion across layers is methodologically orthogonal to spatial fusion across tokens: the former integrates information from different levels of abstraction, while the latter refines localization within a fixed depth.
%This indicates that hierarchical fusion across layers is representationally orthogonal to spatial fusion across tokens: the former integrates abstraction across depth, while the latter refines localization within a fixed depth.
Together, these findings suggest that intermediate-layer fusion is not a replacement for spatial attention, but a complementary axis of representation aggregation that captures information unavailable from patch-level processing alone.

%%%%%%%%%%%%%%%%%%%%   Conclusion   %%%%%%%%%%%%%%%%%%%%%%%%%%%%%%%%%%%%%%%%%%%%%%%%%%%%%%%%%%%%

\section{Discussion}

The field has long hold the belief that most, if not all, task-relevant information is encoded in the last layers of a neural network model~\citep[cf.][]{devlin-etal-2019-bert, zhai2020largescalestudyrepresentationlearning, dosovitskiy2020vit, pmlr-v139-radford21a, kornblith2021loss, raghu2021vision} and, hence, gravitated toward using the penultimate or final layer for adapting model representations via linear probing~\citep{DBLP:conf/iclr/AlainB17, kornblith2019betterimagenetmodelstransfer, muttenthaler2023human}. However, there has recently been suggestive evidence that information relevant for successfully deploying a model downstream may be distributed across several tokens and layers~\citep{oquab2023dinov2, tu2023vqt, chen2024context}.

Here, we provide further evidence that intermediate layers in ViTs encode relevant task-specific signals that the final-layer \CLS representation alone does not capture.
ALF assigns significant weights to both the intermediate and last layers, indicating that intermediate representations contribute meaningful information to downstream predictions. The learned attention weights show that specialized domains like medical and satellite imaging rely heavily on information encoded in intermediate layers, whereas natural image tasks focus on last-layer semantics.
We demonstrate that probing via cross-attention, rather than simple affine transformations, effectively leverages intermediate-layer representations, and show that these benefits hold robustly across different attentive probing architectures.

While standard linear probing becomes unstable when naively extended to multiple layers, our attentive probing mechanism consistently provides improvements across 20 datasets. 
%Although attention over all tokens from the last layer can be highly performative on tasks where precise spatial information is required, it proves brittle with high variance across datasets, making intermediate layers with compact summary tokens a more robust choice for reliable improvements, especially if knowledge about the downstream task is limited. 
Although attention over all tokens from the last layer can be highly performative for tasks requiring precise spatial reasoning, this reflects a different aggregation axis than hierarchical fusion across layers.
%This distinction reflects orthogonal design choices: hierarchical aggregation across layers versus spatial aggregation across patches. 
This distinction reflects orthogonal design choices along independent representational axes: hierarchical aggregation across layers versus spatial aggregation across patches.
For models with \CLS-focused pretraining (CLIP, DINOv2, supervised ViTs), ALF's hierarchical fusion via summary tokens (\CLS+\AP) proves sufficient and is more efficient (Appx.~\ref{app:finetuning}).
Supplementary experiments with Masked Autoencoders (Appx.~\ref{app:mae-results}), whose pretraining is not \CLS-based, show that patch-centric models also benefit from hierarchical aggregation, though direct spatial attention (AAT) becomes more advantageous when information remains distributed across individual patches. However, the combination of these orthogonal fusion approaches shows superior performance (\Cref{fig:intermediate_downstream_overview}), suggesting that incorporating information across the different models' layers is a viable and robust approach for improving downstream adaptation.

\paragraph{Limitations.}
ALF's token selection strategy (\CLS+\AP) is optimized for \CLS-focused pretraining; spatial averaging may neglect localization cues critical to tasks requiring precise spatial reasoning. Additionally, its greater expressivity introduces computational and memory overhead compared to using only the final output token and can increase the risk of overfitting, requiring careful regularization. In addition, the spatial averaging used to summarize the remaining tokens may neglect fine-grained spatial details that some tasks require, particularly those that necessitate precise localization, for which patch-level representations may be more suitable.

\paragraph{Outlook.} The findings of this paper are in accordance with similar discoveries in language models, where intermediate layers can outperform final representations~\citep{DBLP:conf/naacl/Liu0BPS19, skean2025layer}. Together, results across vision and language domains suggest that adaptive access to intermediate representations represents a fundamental principle for the successful deployment of foundation models. This principle extends naturally to emerging biological foundation models for sequences~\citep{evo2}, genomics~\citep{Geneformer}, and proteins~\citep{esm}, where specialized tasks may benefit from intermediate representations that final layers abstract away. As foundation models proliferate across domains, principled methods to access their full representational hierarchy could prove increasingly valuable for maximizing their utility.

% \section*{Acknowledgements}

% \textbf{Do not} include acknowledgements in the initial version of the paper submitted for blind review.

\section*{Acknowledgments}
LC is funded by the Hector Fellow Academy and is grateful for their support. 
MM and SN gratefully acknowledge funding from the German Federal Ministry of Education and Research under the grant BIFOLD25B.
% Lukas T
LT thanks the International Max Planck Research School for Intelligent Systems (IMPRS-IS) and the Carl Zeiss Foundation, project ”Certification and Foundations of Safe Machine Learning Systems in Healthcare” for support. 
% Zeynep (to check)
This work was partially funded by the ERC (853489 - DEXIM) and the Alfried Krupp von Bohlen und Halbach Foundation, for which we thank them for their generous support.
LE and LC would like to thank the European Laboratory for Learning and Intelligent Systems (ELLIS) PhD program for support. 
%S. N. is also supported by the European Union’s HORIZON MSCA Doctoral Networks programme project AQTIVATE (101072344).

\section*{Impact Statement}
This paper studies how to reuse pretrained vision foundation models more effectively using improved probing methods. By enabling strong performance without updating backbone weights, our approach can reduce the computational cost and energy consumption associated with full fine-tuning and may broaden access to high-quality models in low-resource settings.
At the same time, like most advances in representation learning, our method is task-agnostic, therefore, we emphasize that deployment should follow established best practices regarding fairness, robustness, privacy, and human oversight.

% \subsubsection*{Reproducibility statement}
% We provide extensive details to ensure reproducibility of our results. The main paper gives an overview of the experimental setup in Sec.~\ref{subsec:exp-setup}, with further implementation details, including feature extraction, training protocols, hyperparameter search, and regularization strategies, provided in Appx.~\ref{app:implementation-details}. Dataset descriptions are given in Appx. Tab.~\ref{tab:datasets_info}. We release our full code at \url{https://anonymous.4open.science/r/attentive-layer-fusion-8EE0} to enable exact replication of our experiments.

\bibliography{references}
\bibliographystyle{icml2026}

%%%%%%%%%%%%%%%%%%%%%%%%%%%%%%%%%%%%%%%%%%%%%%%%%%%%%%%%%%%%%%%%%%%%%%%%%%%%%%%
%%%%%%%%%%%%%%%%%%%%%%%%%%%%%%%%%%%%%%%%%%%%%%%%%%%%%%%%%%%%%%%%%%%%%%%%%%%%%%%
% APPENDIX
%%%%%%%%%%%%%%%%%%%%%%%%%%%%%%%%%%%%%%%%%%%%%%%%%%%%%%%%%%%%%%%%%%%%%%%%%%%%%%%
%%%%%%%%%%%%%%%%%%%%%%%%%%%%%%%%%%%%%%%%%%%%%%%%%%%%%%%%%%%%%%%%%%%%%%%%%%%%%%%
\newpage
\appendix
\onecolumn
\section{Appendix}

\subsection{Implementation Details} \label{app:implementation-details}
This section describes the technical implementation approach and experimental setup used to evaluate the attention-based intermediate layer fusion mechanisms.

A frozen backbone strategy was adopted, training only the attention fusion mechanism and classification head on top of pre-extracted features. The latent representations (of intermediate and last layers) for each model-dataset combination were extracted using the Python package \texttt{thingsvision}~\citep{Muttenthaler_2021}, and the experiment code was built on top of the code from \citet{ciernik2025objective}. Input images were resized to 256px and center-cropped to 224px before applying the model-specific normalizations from the pre-training. Extracted features were then L2-normalized to yield comparable magnitudes. To handle models with varying feature dimensions across layers (e.g., CLIP), we ensured dimensional consistency through zero-padding.

All models were trained for at least 40 epochs using AdamW optimization with cosine annealing learning rate scheduling and a batch size of at most 2048. For small datasets, we adjusted the batch sizes to ensure at least 5 batches per epoch, and increased the number of epochs to guarantee at least 1000 gradient update steps.

To address class imbalance, we trained with a weighted cross-entropy objective~\citep{aurelio_learning_2019}, scaling each class by the inverse of its frequency. The loss is $\text{Loss}(y, \hat{y}) = -\frac{1}{N} \sum_{j=1}^{N} \sum_{i=1}^{K} w_i y_{ji} \log \hat{y}_{ji} $, where $y_{ji}$ is the one-hot ground-truth label for sample $j$ and class $i$, and $\hat{y}_{ji}$ is the predicted probability. $w_i$ are class weights computed as $w_i = \frac{N}{K \cdot n_i}$, with $N$ being the total number of training samples, $K$ the number of classes, and $n_i$ the number of samples in class $i$. This weighting balances learning across minority and majority classes.

Hyperparameter selection used a stratified 80/20 train-validation split with grid search over learning rates $\{0.1, 0.01, 0.001\}$, attention dropout rates $\{0.0, 0.1, 0.3\}$, and weight decay values $\{10^{-6}, 10^{-5}, 10^{-4}, 0.001, 0.01, 0.1, 1.0\}$, except for the AAT baseline, where we used the reported weight decay of 0.1~\cite{chen2024context}. We selected the combination that achieved the best validation balanced accuracy.

To prevent overfitting, we applied gradient norm clipping at 5.0 and added Gaussian noise $\mathcal{N}(0, 0.05)$ to representations with probability 0.5 during training.

For the representation-fusion attention mechanism, we adjust the number of heads to match the number of representations being fused (cf. Appx.~\ref{app:nr-heads}). For example, when fusing \CLS and \AP tokens from all 12 layers of a ViT-B-16 model, we used $M=24$ heads. For the AAT baseline, we used 8 attention heads following~\cite{chen2024context}, as increasing the number of heads did not yield substantial improvements. The learned query tokens were initialized from a normal distribution $\mathcal{N}(0, 0.02)$.

\subsection{Model Details}
\label{app:model-details}
This section provides the specific model variants and patch sizes used in our experiments across three model families: supervised ViTs, self-supervised DINOv2 models, and image-text aligned CLIP models.

\begin{itemize}
    \item \textbf{Supervised ViT:} ViT-S/16, ViT-B/16, and ViT-L/16 pretrained on ImageNet-21K and fine-tuned on ImageNet-1K ~\citep{imagenet1kpaper,imagenet21kpaper}.
    \item \textbf{Self-Supervised DINOv2:} ViT-S-14, ViT-B-14, and ViT-L-14 , pretrained on the LVD-142M dataset~\citep{oquab2023dinov2}.
    \item \textbf{Image-Text Alignment CLIP:} OpenCLIP models ViT-B-32, ViT-B-16, and ViT-L-14~\citep{Cherti_2023_CVPR,ilharco_gabriel_2021_5143773}) following the CLIP architecture and using its pretrained weights~\citep{pmlr-v139-radford21a}). As a small-capacity CLIP model, we use ViT-B/32; its larger patch size significantly reduces the number of input tokens, making its computational and representational capacity analogous to the ``Small'' variants in the other families.
\end{itemize}

\subsection{Dataset Details}
\begin{table}[ht]
\centering
\caption{Overview of the 20 datasets used in our experiments including the size of both train and test set, number of classes, and the Class Imbalance Ratio (CIR) calculated by $\frac{N_{\text{Majority Class}}}{N_{\text{Minority Class}}}$.}
\resizebox{\linewidth}{!}{
\begin{tabular}{llrrrrl}
\hline
Category & Dataset & Train Size & Test Size & Classes & CIR & Reference \\
\hline
\multirow{6}{*}[0.5ex]{Natural (MD)}  & STL-10 & 5\,000 & 8\,000 & 10 & 1 & \citet{pmlr-v15-coates11a} \\
& CIFAR-10 & 45\,000 & 10\,000& 10 & 1.02 & \citet{krizhevsky2009learning} \\
& Caltech-101 & 2\,753 & 6\,085 & 102 & 1.3 & \citet{li2006caltech} \\
& PASCAL VOC 2007 & 7\,844 & 14\,976 & 20 & 20.65 & \citet{DBLP:journals/ijcv/EveringhamGWWZ10} \\
& ImageNet-1k & 1\,281\,167  & 50\,000 & 1000  & 1.78 & \citet{imagenet1kpaper} \\
& CIFAR-100 & 45\,000 & 10\,000 & 100 & 1.06 & \citet{krizhevsky2009learning} \\
& Country-211 & 31\,650 & 21\,100 & 211 & 1 & \citet{pmlr-v139-radford21a} \\
\hline
\multirow{6}{*}[0.5ex]{Natural (SD)} & Pets & 2\,944 & 3\,669 & 37 & 1.24 & \citet{DBLP:conf/cvpr/ParkhiVZJ12} \\
& Flowers & 1\,020 & 6\,149 & 102 & 1 & \citet{DBLP:conf/icvgip/NilsbackZ08} \\
& Stanford Cars & 8\,144 & 8\,041 & 196 & 2.83 & \citet{DBLP:conf/iccvw/Krause0DF13} \\
& FGVC Aircraft & 3\,334 & 3\,333 & 100 & 1.03 & \citet{maji2013fine} \\
& GTSRB & 26\,640 & 12\,630 & 43 & 10 & \citet{DBLP:journals/nn/StallkampSSI12} \\
& SVHN & 65\,931 & 26\,032 & 10 & 2.98 & \citet{netzer2011svhn} \\
\hline
\multirow{4}{*}[0.5ex]{Specialized} & PCAM & 262\,144 & 32\,768 & 2 & 1 & \citet{DBLP:conf/miccai/VeelingLWCW18} \\
& EuroSAT & 16\,200 & 5\,400 & 10 & 1.58 & \citet{DBLP:journals/staeors/HelberBDB19} \\
& RESISC45 & 18\,900 & 6\,300 & 45 & 1.16 & \citet{DBLP:journals/pieee/ChengHL17} \\
& Diabetic Retinopathy & 35\,126 & 42\,670 & 5 & 36.45 & \citet{diabetic-retinopathy-detection} \\
\hline
\multirow{3}{*}[0.5ex]{Structured} & DTD & 1\,880 & 1\,880 & 47 & 1 & \citet{cimpoi2014describing} \\
& FER2013 & 28\,709 & 7\,178 & 7 & 16.55 & \citet{goodfellow2013challenges} \\
& Dmlab & 65\,550 & 22\,735 & 6 & 1.98 & \citet{zhai2020largescalestudyrepresentationlearning} \\
\hline
\end{tabular}
}
\label{tab:datasets_info}
\end{table}
An overview of all datasets used in this work is given in \Cref{tab:datasets_info}. Following VTAB \citep{zhai2020largescalestudyrepresentationlearning}, the datasets are categorized by domain. We separate natural images into multi-domain (MD) and single-domain (SD) datasets, and include specialized as well as structured datasets.

\subsection{Downstream performance for all datasets and models}
\label{app:single_model_complete_results}
We present the balanced test accuracies across all 20 downstream datasets for each of our nine models. Each table (Figures~\ref{fig:perf_clip_comparison},~\ref{fig:perf_dinov2_comparison}, and~\ref{fig:perf_vit_comparison}) shows four different probing configurations: (1) last layer \CLS token with linear probing, (2) last layer all tokens with attentive probing, (3) all layers \CLS and \AP token with linear probing, and (4) all layers \CLS and \AP token with attentive probing. 

The bottom rows of each table report summary statistics of the absolute performance gains relative to the baseline last-layer CLS linear probe, including the minimum, median, maximum, mean, and standard deviation of improvements across all datasets. Color coding indicates relative performance within each model family, with darker colors representing better performance.
% \begin{figure}[H]
%     %\vspace{0em}
%     \centering
%     \includegraphics[width=0.97\linewidth]{tables/single_model_tables/clip.png}
%     \vspace{-0.5em} 
%     \caption{From left to right: \textbf{CLIP-B-32} (small), \textbf{CLIP-B-16 }(base), \textbf{CLIP-L-14} (large).}
%     \label{fig:perf_clip_comparison}
%     %\vspace{-4em} 
    
% \end{figure}
% \begin{figure}[H]
%     \vspace{-13em}
%     \centering
%     \includegraphics[width=0.97\linewidth]{tables/single_model_tables/dinov2.png}
%     \vspace{-0.5em} 
%     \caption{From left to right: \textbf{DINOv2-S-14} (small), \textbf{DINOv2-B-14} (base), \textbf{DINOv2-L-14} (large).}
%     \label{fig:perf_dinov2_comparison}
%     %\vspace{-4em} 
    
% \end{figure}
% \begin{figure}[H]
%     \vspace{-13em}
%     \centering
%     \includegraphics[width=0.97\linewidth]{tables/single_model_tables/vit.png}
%     \vspace{-0.5em} 
%     \caption{From left to right: \textbf{ViT-S-16} (small), \textbf{ViT-S-16} (base), \textbf{ViT-L-16} (large).}
%     %\vspace{-4em} 
%     \label{fig:perf_vit_comparison}
% \end{figure}

\begin{figure}
    \begin{subfigure}{\textwidth}
        \centering
        \includegraphics[width=0.8\linewidth]{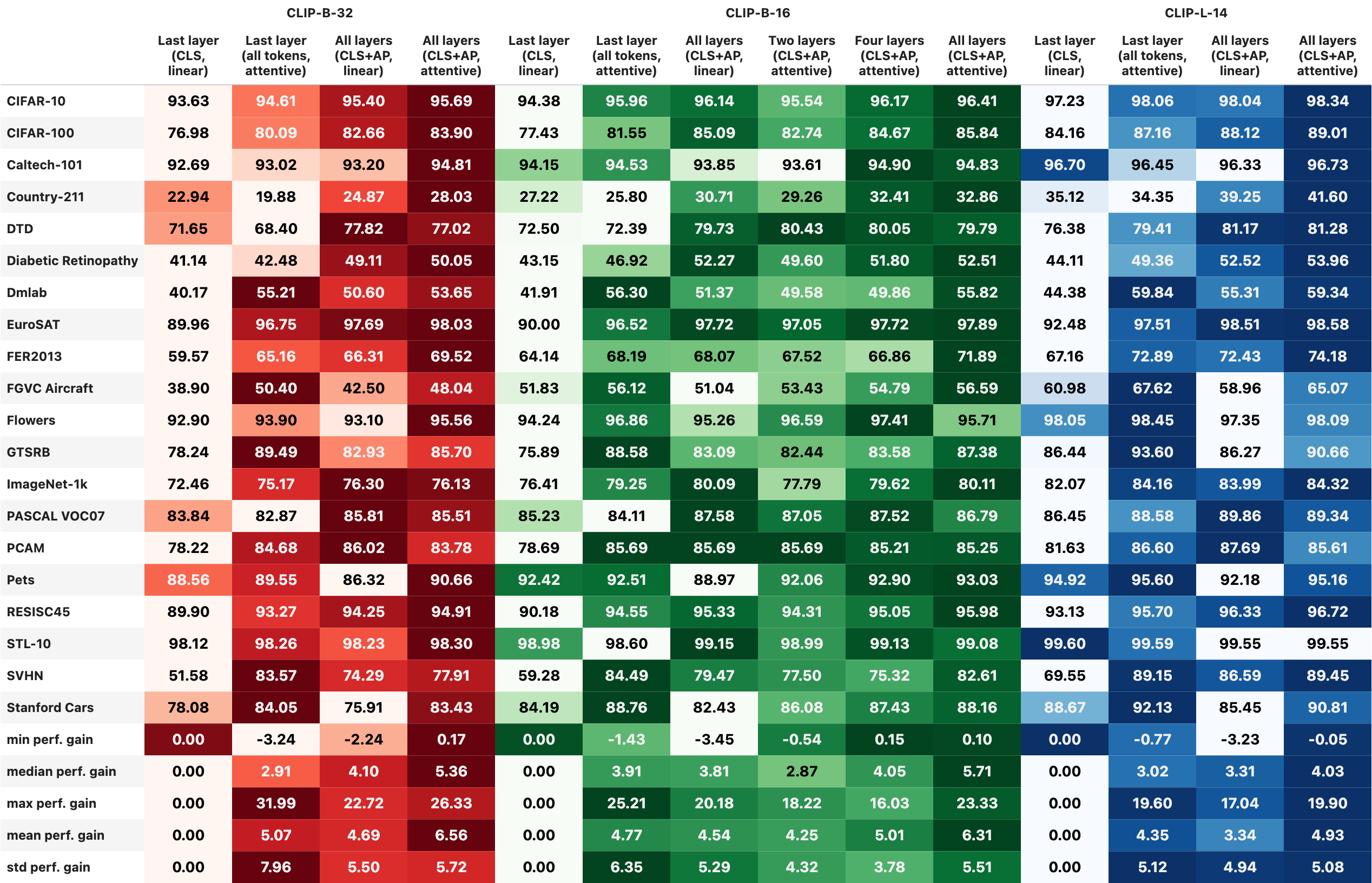}
        \caption{From left to right: \textbf{CLIP-B-32} (small), \textbf{CLIP-B-16} (base), \textbf{CLIP-L-14} (large).}
        \label{fig:perf_clip_comparison}
    \end{subfigure}
    \begin{subfigure}{\textwidth}
        \centering
        \includegraphics[width=0.8\linewidth]{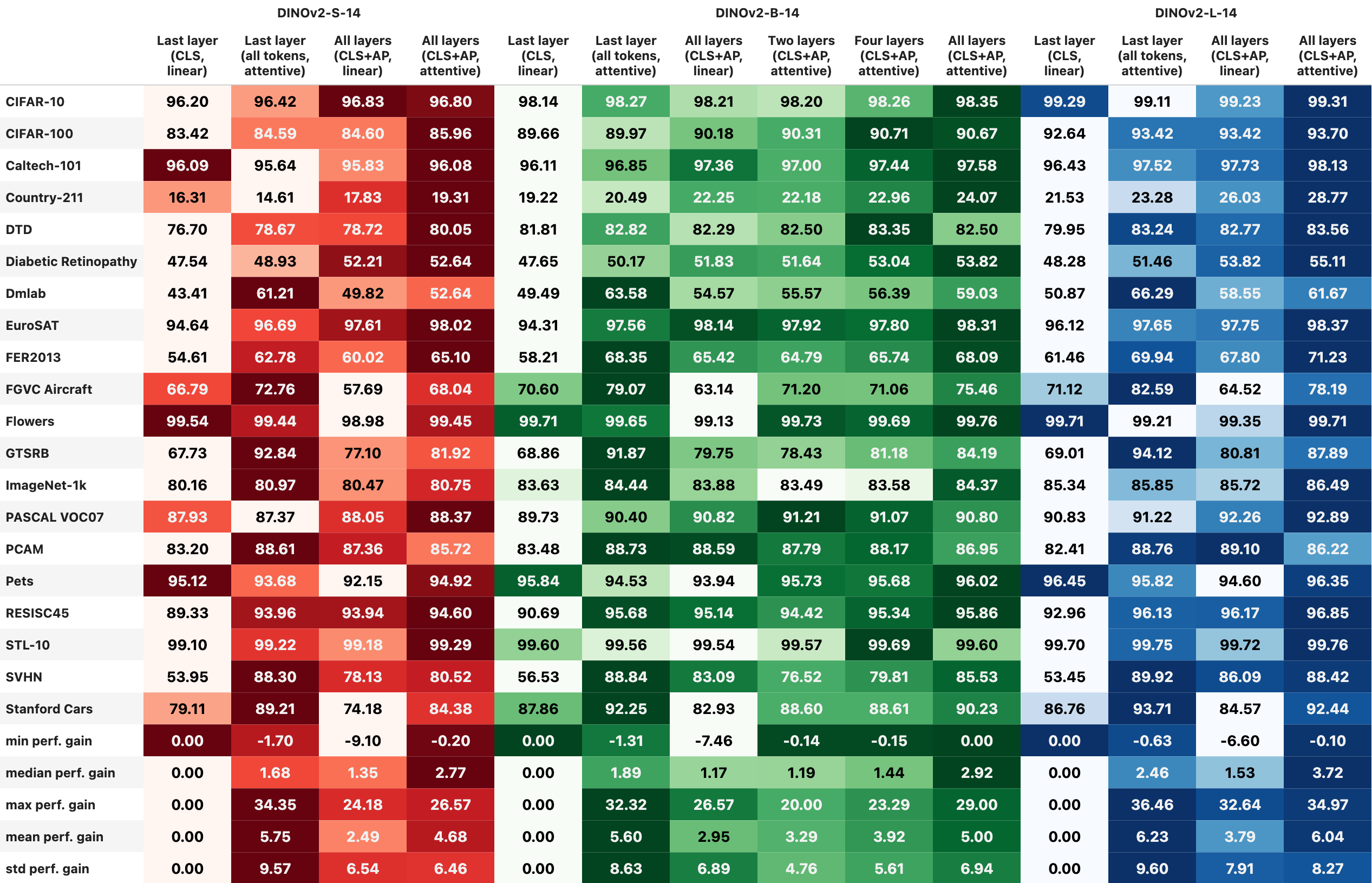}
        \caption{From left to right: \textbf{DINOv2-S-14} (small), \textbf{DINOv2-B-14} (base), \textbf{DINOv2-L-14} (large).}
        \label{fig:perf_dinov2_comparison}
    \end{subfigure}
    \caption{Accuracies per model and dataset}
    \label{fig:all_models_detailed_perfs}
\end{figure}
\begin{figure}[ht]\ContinuedFloat
    \begin{subfigure}{\textwidth}
        \centering
        \includegraphics[width=0.8\linewidth]{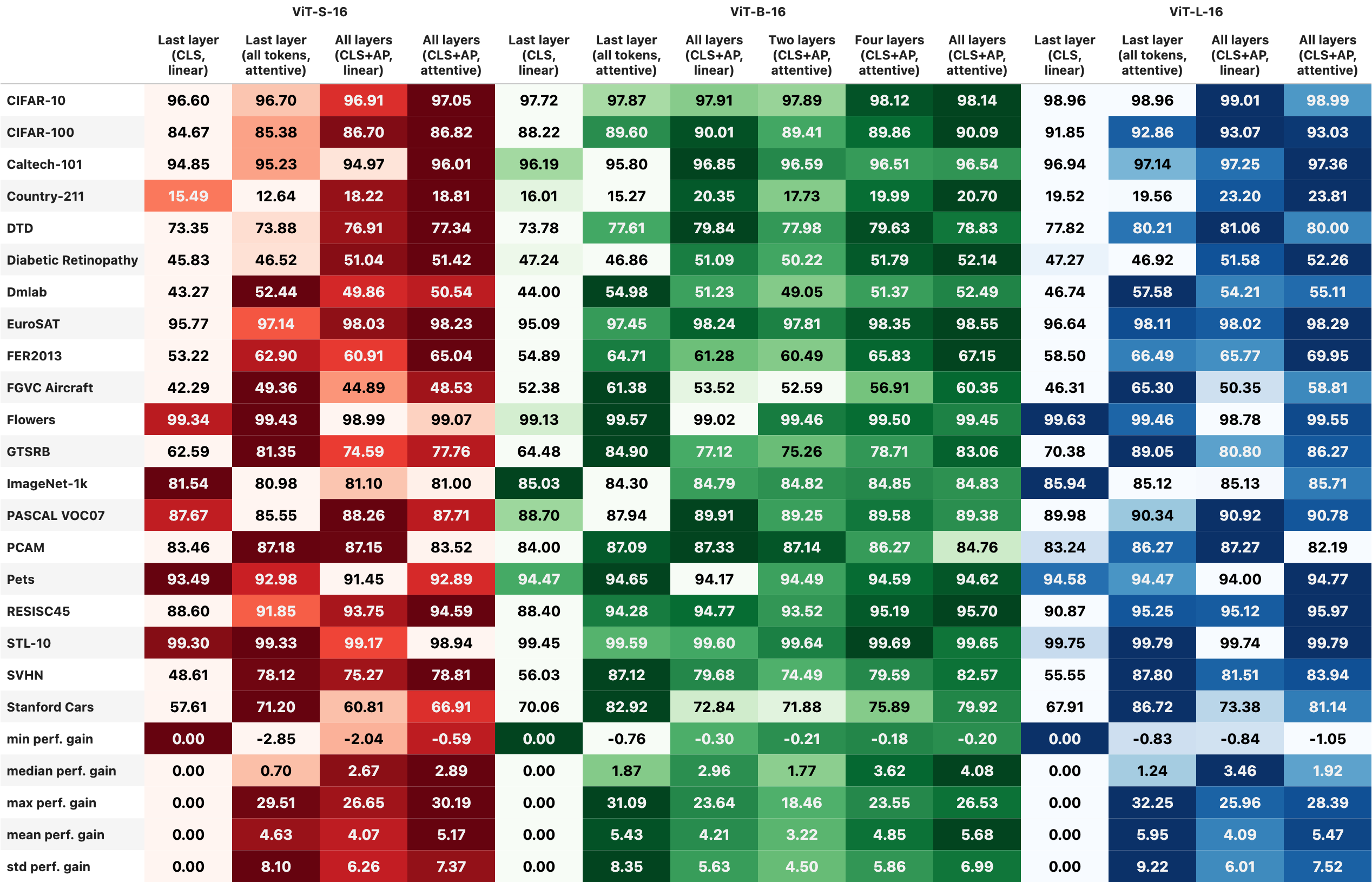}
        \caption{From left to right: \textbf{ViT-S-16} (small), \textbf{ViT-B-16} (base), \textbf{ViT-L-16} (large).}
        \label{fig:perf_vit_comparison}
    \end{subfigure}
    \caption{Accuracies per model and dataset}
    \label{fig:all_models_detailed_perfs}
\end{figure}

\subsection{Attention Heatmaps for small and large models}\label{app:small_large_attn_heatmaps}
\Cref{fig:appx:attn_smalllarge} compares the aggregated attention across small and large models. Despite substantial differences in scale and twice as many layers for the large models,  the attention patterns are very similar. This underlies our intuition that the relevance of intermediate layers depends more on the task characteristics than on model size or objective, which seem to learn very similar hierarchies. Specialized and structural datasets drive attention toward intermediate layers, while natural image datasets close to the pre-training domain rely more on the later-layer \CLS tokens. Notably, in cases like GTSRB and SVHN, where linear \CLS probing fails but ALF achieves large gains, the probe shifts attention to the \AP tokens. These results reinforce that ALF adapts flexibly to task demands while remaining consistent across models of very different scales and pre-training objectives.
\begin{figure}[H]
\begin{subfigure}{0.95\textwidth}
    \centering
    \includegraphics[width=1\linewidth]{figures/appendix/attention_heatmaps/AttentionWeightsSmallModels.png}
    \label{fig:appx:attn_small}
\end{subfigure}
\begin{subfigure}{0.95\textwidth}
    \centering
    \includegraphics[width=1\linewidth]{figures/appendix/attention_heatmaps/AttentionWeightsLargeModels.png}
    \label{fig:appx:attn_large}
\end{subfigure}
    \caption{Aggregated attention maps from ALF for small (top) and large (bottom) models. Attention patterns vary more across datasets than across model scales, underscoring the task-dependent relevance of intermediate-layer features.}
    \label{fig:appx:attn_smalllarge}
\end{figure}

\subsection{Layer Pruning Analysis}
\label{app:layerpruning}
\begin{figure}[t]
    \centering
    \includegraphics[width=\linewidth]{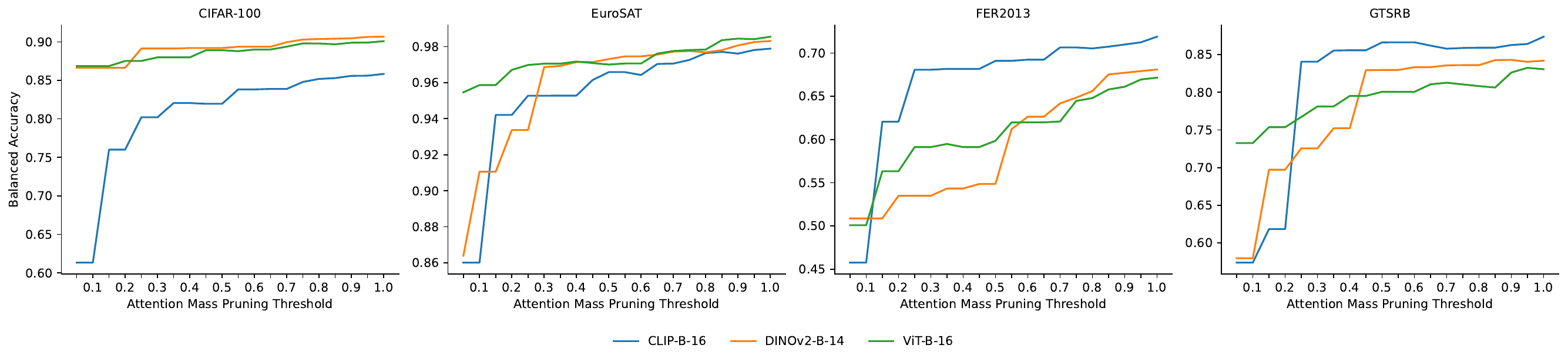}
    \caption{Performance under cumulative attention-mass pruning. Tokens are retained according to the top-$x\%$ of total attention mass computed from population-level averaged attention weights.}
    \label{fig:attn_mass_pruning}
\end{figure}

\begin{figure}[t]
    \centering
    \includegraphics[width=\linewidth]{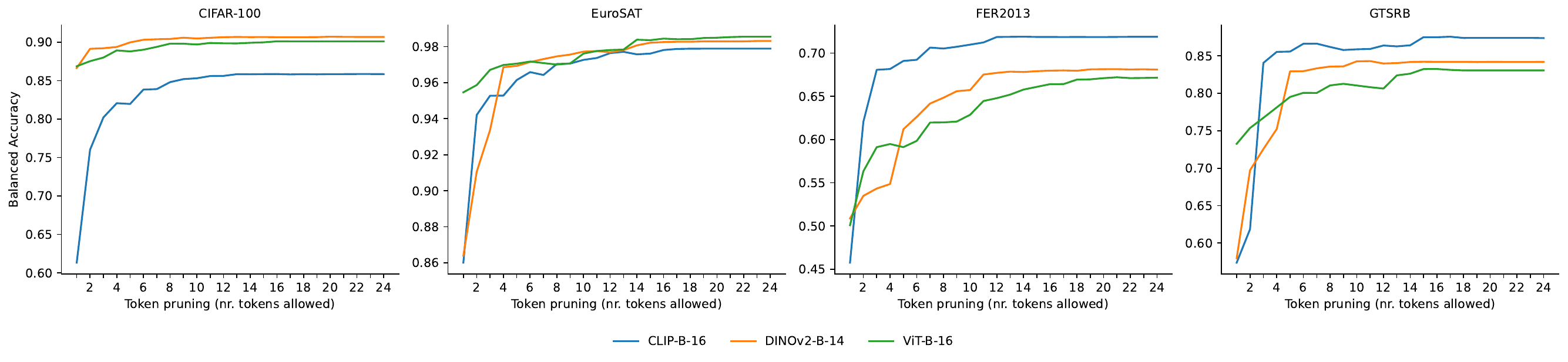}
    \caption{Performance when retaining only the top-$N$ highest-attended summary tokens.}
    \label{fig:topn_pruning}
\end{figure}
To further analyze the sparsity induced by ALF attention, we investigate whether aggregating representations from all layers is necessary by evaluating two inference-time pruning strategies based on learned attention weights.
We first compute population-level attention weights by averaging attention scores across all test samples and all 24 heads, then prune tokens according to their cumulative attention mass. Specifically, we retain only the subset of tokens accounting for the top-$x\%$ of total attention mass and mask the rest.
Fig.~\ref{fig:attn_mass_pruning} shows that performance generally improves as more attention mass is retained across datasets and backbones. Moderate pruning (retaining approximately 60--70\% of attention mass) yields only minor degradation, indicating that many low-attention tokens can be removed with limited loss. However, no pruning configuration matches the full performance, suggesting that even low-weight tokens provide complementary information.
We additionally prune by retaining only the top-$N$ highest-attended summary tokens, directly testing whether a reduced subset of layers is sufficient.
As shown in Fig.~\ref{fig:topn_pruning}, near-full performance is recovered using roughly half of the available tokens in most settings, confirming that ALF already performs effective soft selection over layers. 

For both pruning strategies, model sensitivity differs across architectures and datasets, confirming that the optimal subset of informative layers is model- and task-dependent. Therefore, a fixed manual layer selection is less robust than extracting all layers and allowing learned attention to perform adaptive routing.

\subsection{Additional Experiments with Masked Auto Encoder}\label{app:mae-results}
\begin{figure}[t]
    \centering
    \includegraphics[width=\linewidth]{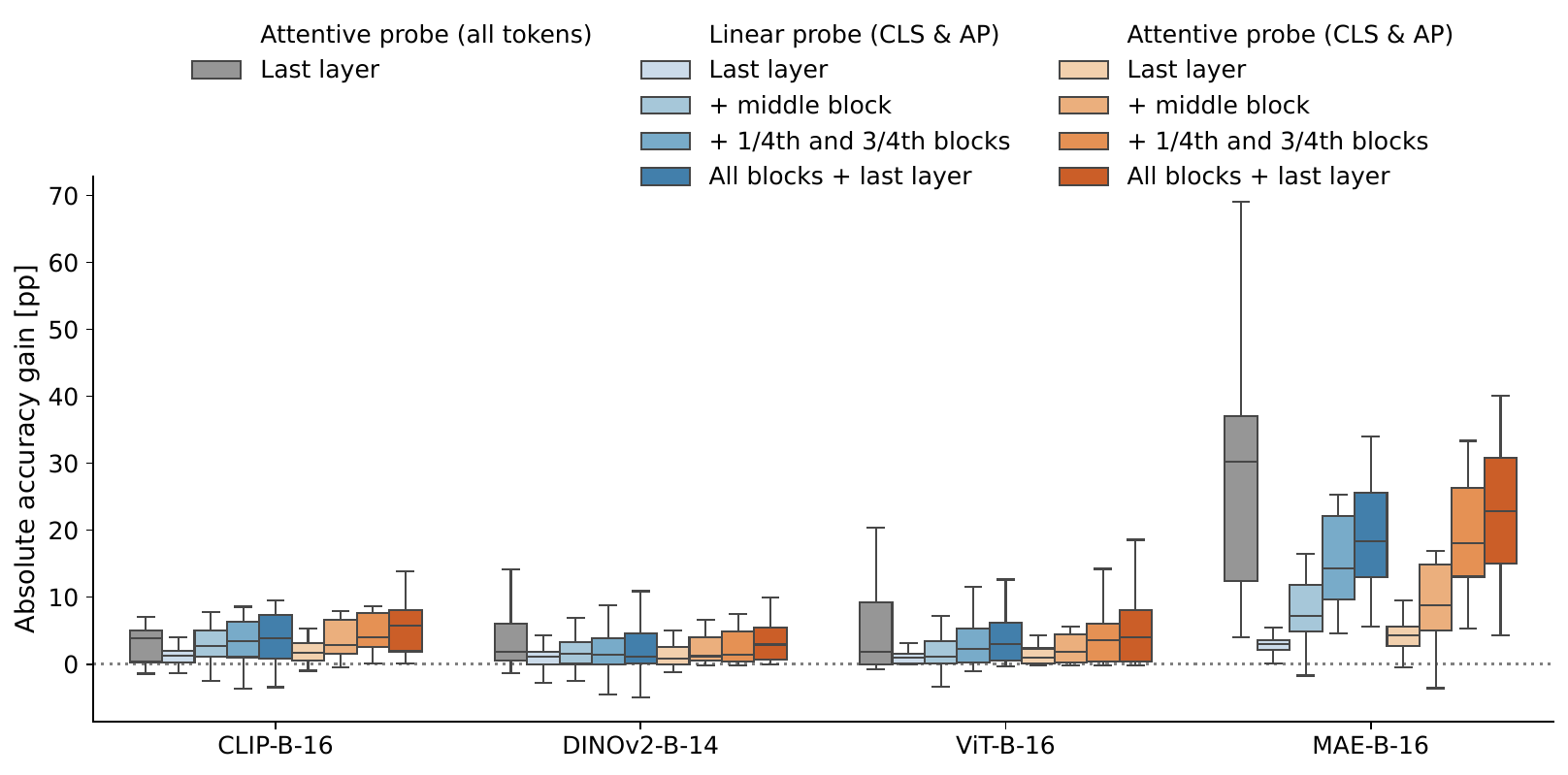}
    \caption{
    Absolute accuracy gain (percentage points) of linear (blue) and attentive probes (orange) when fusing an increasing number of intermediate layer representations (\lastlayer, \twolayers, \fourlayers, and \alllayers), as well as AAT (grey) aggregated across datasets for the three base models as well as the \textbf{base MAE model}. For MAE, the simple linear probe on \AP tokens is insufficient for most tasks, explaining the large gain in accuracy by either including spatial (all tokens last layer) or hierarchical (\CLS \& \AP, all blocks) information.
    }
    \label{fig:perf_gain_boxplots_base_models_with_mae}
\end{figure}

\begin{figure}
    \centering
    \includegraphics[width=0.8\linewidth]{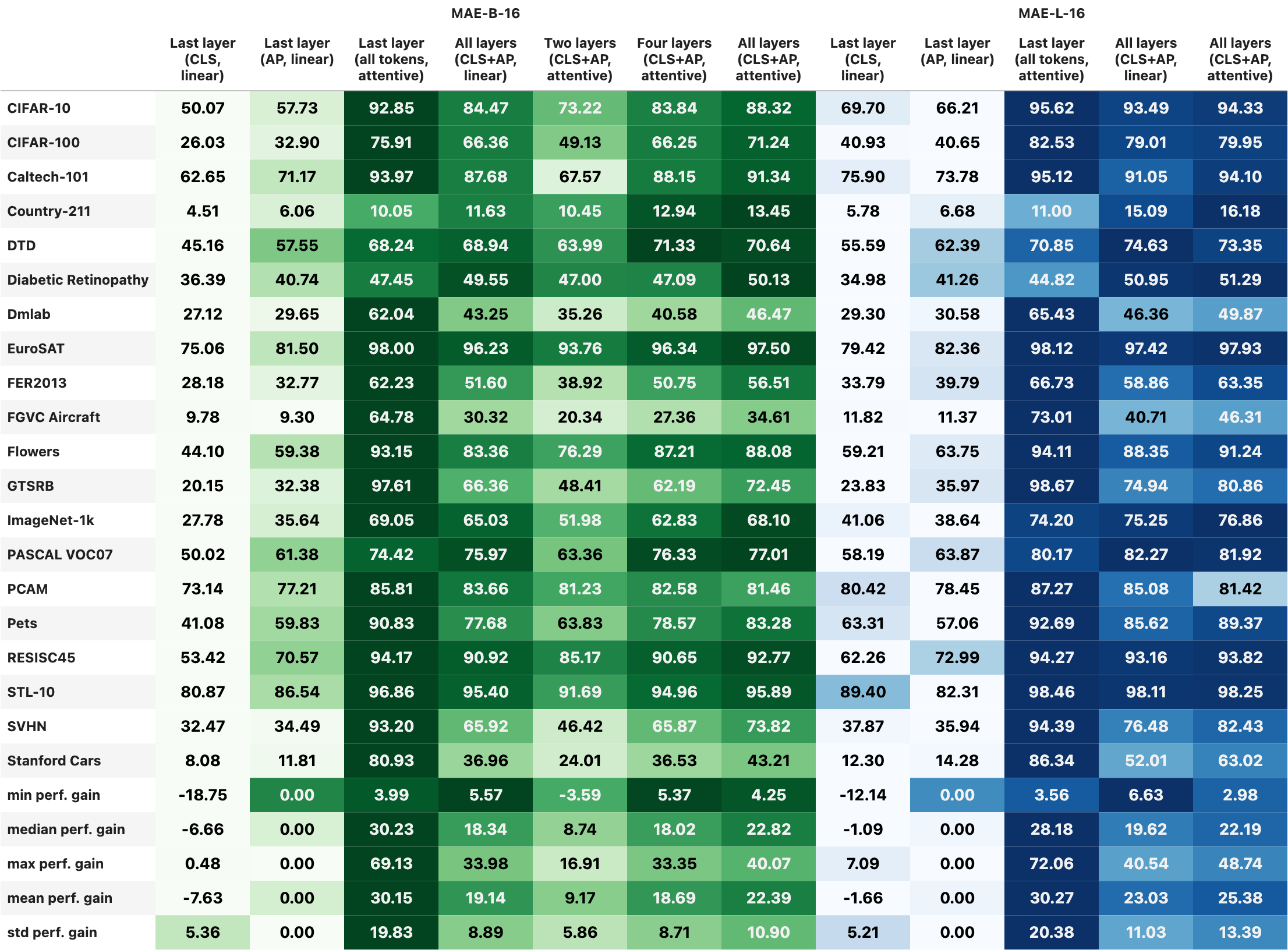}
    \caption{Detailed results of both base and large MAE on all datasets. While attending over intermediate layer provides already a large benefit, the aggregation over all tokens is a necessity for masked image modeling confirming that the information is distributed over image tokens. }
    \label{fig:maefullresults}
\end{figure}

Masked Autoencoders (MAEs) \citep{DBLP:conf/cvpr/HeCXLDG22} represent a distinct class of pretrained models whose representational structure differs fundamentally from that of CLIP, DINOv2, or supervised ViTs, as they are trained exclusively via patch-level reconstruction and thus do not use summary tokens in their loss. Prior work \citep{przewikezlikowski2025beyond} has shown that MAEs retain highly localized information until the final layers and therefore benefit most from probes that attend over all patch tokens rather than relying on summary tokens. We confirm this observation: for both MAE-B-16 and MAE-L-16, an attentive probe operating over all tokens achieves the strongest overall performance (\Cref{fig:perf_gain_boxplots_base_models_with_mae} and \Cref{fig:maefullresults}).
Nevertheless, ALF, which operates only on the aggregated \CLS/\AP tokens, still produces large gains over last-layer \AP probes, with mean improvements of 22.4 (base) or 24.7 (large) percentage points. In most datasets, it ranks second only to the full-token attentive probe, and on two datasets, it even surpasses it. This demonstrates that layer-wise fusion recovers complementary information across depth, highlighting the orthogonality between token aggregation (spatial dimension) and layer aggregation (hierarchical dimension).
These results illustrate an important representational difference. MAEs do not compress information into the \CLS token, so probes that access all patch tokens are inherently favored. Nevertheless, when restricted to summary tokens, multi-layer fusion substantially mitigates this limitation. Thus, our results on MAE confirm that performance depends both on how information is distributed across tokens and how it evolves across layers. ALF remains effective, especially when a model exposes meaningful layerwise summary representations.

To better understand these results, we inspect the learned attention patterns for MAE-B-16 and MAE-L-16 (\Cref{fig:maeattention}). Since the MAE \CLS token is not trained, the probe naturally places nearly all its attention on the \AP tokens, confirming that summary representations are weak in this model family.
The attention also concentrates on the later layers, consistent with the fact that MAEs preserve spatial detail until the end of the network and perform little semantic compression. As a result, aggregating information from intermediate \AP tokens enables ALF to recover much of the depth-wise structure, allowing it to approach the performance of probes with access to all spatial tokens.
\begin{figure}
    \centering
    \includegraphics[width=0.98\linewidth]{figures/rebuttal/MAEAttentionMapOverDatasetPerModelswoIN.png}
    \caption{Aggregated intermediate-layer attention maps for MAE-B-16 and MAE-L-16 show that MAEs store task-relevant information predominantly in later \AP tokens.}
    \label{fig:maeattention}
\end{figure}

\subsection{Relationship between intermediate-layer performance and representational similarity}\label{app:intermediate-layer-alone}
Prior work has shown that intermediate layers contain task-relevant information accessible via linear probing~\citep{DBLP:conf/iclr/AlainB17}. Following \citet{pmlr-v97-kornblith19a}, we examine the relationship between downstream performance and representational similarity measured by Centered Kernel Alignment (CKA) with RBF kernel ($\sigma = 0.2$), which emphasizes local neighborhood similarities relative to the final layer's representation. To study this across architectures and feature types, we trained linear probes on all intermediate layers of the four base models (CLIP-B-16, DINOv2-B-14, ViT-B-16, MAE-B-16) on CIFAR-100, GTSRB, FER2013, and EuroSAT. For all models except MAE, we probe the \CLS token, while for MAE, we use the \AP token.

\begin{figure}
    \centering
    \includegraphics[width=\linewidth]{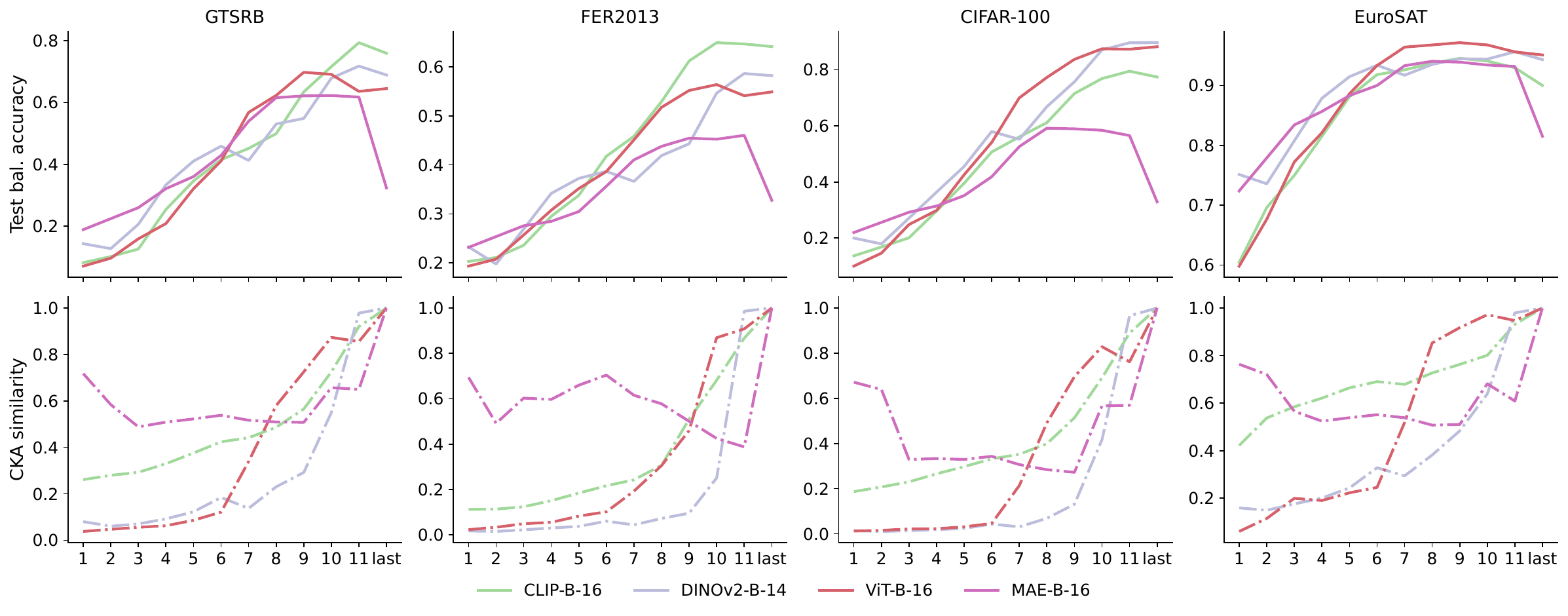}
    \caption{Downstream performance vs. representational similarity across intermediate layers. \textbf{Top row}: test balanced accuracy of linear probes on layers 1-11 and the final layer. \textbf{Bottom row}: CKA similarity between each layer and the final layer. Intermediate layers can achieve high performance despite low similarity to the final layer.}
    \label{fig:cka_local_vs_downstream_perf}
\end{figure}

\Cref{fig:cka_local_vs_downstream_perf} shows that CKA similarity to the final layer is not strongly predictive of downstream performance. While similarity tends to increase rapidly in the later layers, the largest accuracy gains often occur in early or middle layers. Notably, even though these intermediate representations are dissimilar to the final layer, they achieve similar or higher performance, for datasets like GTSRB and EuroSAT, the performance even peaks at layer 6-8.

These results suggest that intermediate layers capture complementary features that are not redundant with the final-layer representations, motivating adaptive fusion strategies to leverage this diverse information effectively.

\subsection{Comparing per-layer linear and attentive probe performance}\label{app:intermediate-layer-performance}
\begin{figure}
    \centering
    \includegraphics[width=\linewidth]{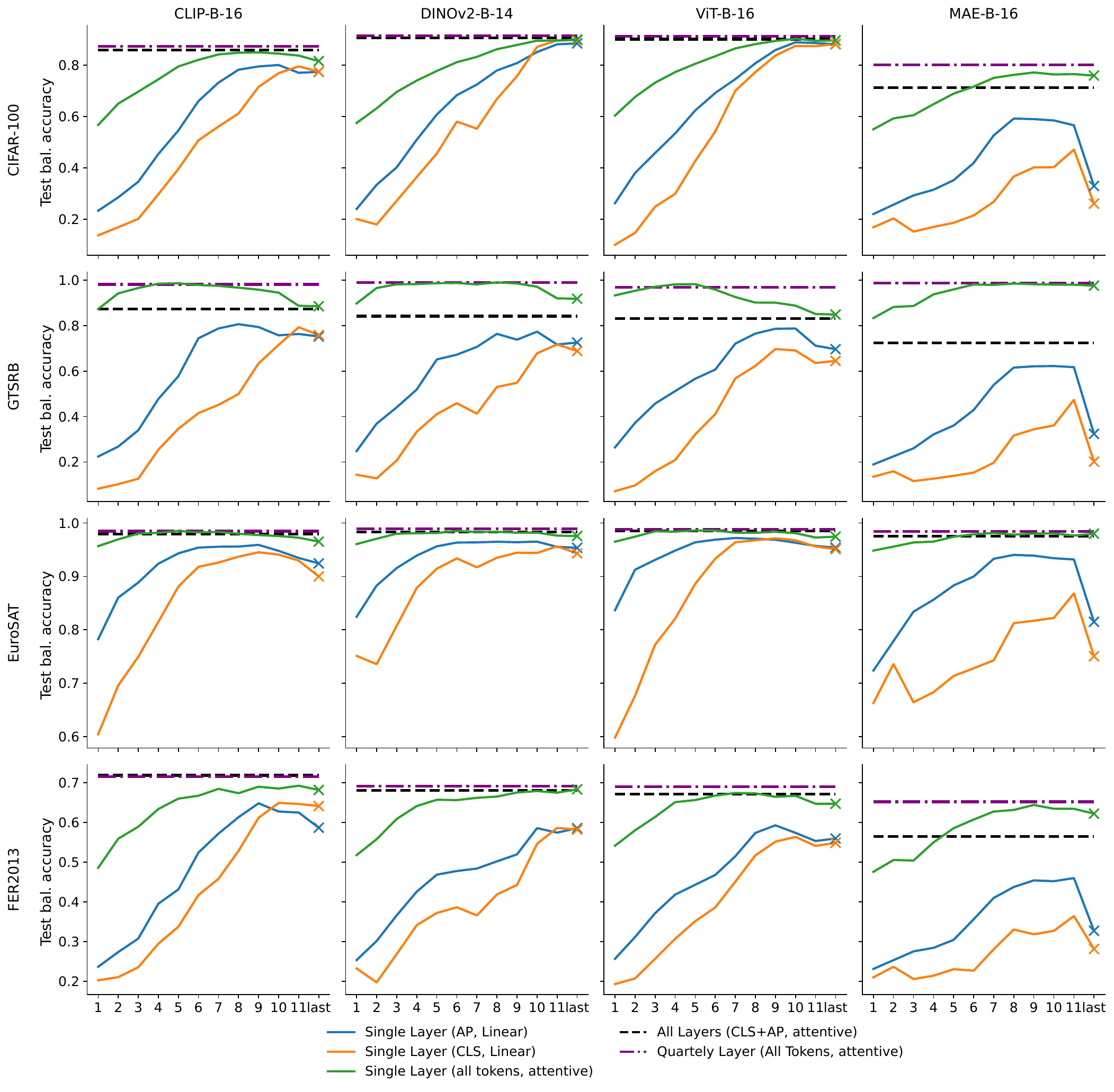}
    \caption{
    %Downstream performance across intermediate layers for linear probe with \AP and \CLS token and attentive probe on all tokens. The dashed line indicates ALF, which aggregates only the \CLS and \AP tokens across layers. Additionally, the purple dash-dotted line shows a hybrid approach, aggregating all tokens of intermediate layers.
    Downstream performance across intermediate layers. Comparison between linear probing (using \AP and \CLS tokens) and attentive probing across all tokens. The black dashed line represents ALF, which aggregates only \CLS and \AP tokens across layers, while the purple dash-dotted line denotes a hybrid approach that aggregates all layer tokens from quarterly-separated layers. 
    }
    \label{fig:intermediate_downstream_overview}
\end{figure}

In this section, we contrast three per-layer probing strategies, linear probing on the \CLS and \AP tokens, and an attentive probe that aggregates all tokens of a layer, with our multi-layer attentive fusion, which operates only on the aggregated \CLS and \AP representations. Additionally, we add a specialized hybrid probe as discussed below. Results for four datasets and the base models are shown in \Cref{fig:intermediate_downstream_overview}.

Across all settings, the per-layer attentive probe substantially outperforms linear probes on \CLS or \AP tokens, indicating that intermediate-layer information is distributed across spatial tokens and cannot entirely be recovered from a single summary embedding. Despite operating under the stricter constraint of using only \CLS and \AP tokens, our multi-layer fusion often exceeds the best per-layer attentive probe. By combining complementary information across depth, our multi-layer attentive fusion recovers much of the signal lost in token aggregation.

Two cases deviate from this trend. For GTSRB, performance peaks in shallow layers and is driven by highly localized features that are not preserved in \CLS or \AP tokens, making full-token attention inherently stronger.
For MAE, patch tokens encode rich, localized structure from reconstruction training, whereas the \CLS token receives no explicit supervision to serve as a global summary.
Thus, average pooling discards information that an attentive probe over all tokens can utilize. These behaviors are expected given the architectural differences and highlight that, even under strong token-aggregation constraints, ALF consistently outperforms all its per-layer components by combining complementary information across layers.
To further validate this orthogonality between hierarchical and spatial aggregation, we introduce a hybrid attention probe that combines all tokens from layers $3,6,9$ and the last layer. To stabilize training with this larger token set ($788$ for CLIP, ViT, MAE, and $1028$ for DINOv2), we increase attention dropout to $0.5$ and the number of heads to $24$, leaving all other hyperparameters untouched. This hybrid probe consistently outperforms both AAT applied to the last layer and ALF relying only on summary tokens. The magnitude of improvement depends on the dataset: gains are modest when summary tokens suffice (e.g., CIFAR-100, EuroSAT), but substantial when spatial details are essential (e.g., GTSRB) or when the backbone distributes information across patch tokens, as in MAE. These results further support our claim that intermediate-layer fusion and patch-token selection operate on orthogonal representational axes and that incorporating intermediate-layer tokens is significantly more effective than relying solely on the last layer.

\subsection{Parameter efficiency comparison}\label{appx:parameter_count}
\begin{table}[htp]
\centering
%\footnotesize
\caption{Parameter counts for Linear and Attentive fusion probes using all layers ({\CLS}+{\AP}) across different numbers of classes ($K$) and three ViT architectures: \\ ViT-S ($d=384$, $|\alllayers|=12$), ViT-B ($d=768$, $|\alllayers|=12$), and ViT-L ($d=1024$, $|\alllayers|=24$).}
%\resizebox{\linewidth}{!}{
\begin{tabular}{r|rr|rr|rr}
\hline
  & \multicolumn{2}{c|}{$d=384$, $|\alllayers|=12$} & \multicolumn{2}{c|}{$d=768$, $|\alllayers|=12$} & \multicolumn{2}{c}{ $d=1024$, $|\alllayers|=24$} \\
$K$ & Linear & Attentive & Linear & Attentive & Linear & Attentive \\
\hline
2   & 18\,434    & 1\,184\,258 & 36\,866    & 4\,727\,810 & 98\,306     & 8\,400\,898 \\
5   & 46\,085    & 1\,185\,413 & 92\,165    & 4\,730\,117 & 245\,765    & 8\,403\,973 \\
10  & 92\,170    & 1\,187\,338 & 184\,330   & 4\,733\,962 & 491\,530    & 8\,409\,098 \\
50  & 460\,850   & 1\,202\,738 & 921\,650   & 4\,764\,722 & 2\,457\,650  & 8\,450\,098 \\
100 & 921\,700   & 1\,221\,988 & 1\,843\,300 & 4\,803\,172 & 4\,915\,300  & 8\,501\,348 \\
200 & 1\,843\,400 & 1\,260\,488 & 3\,686\,600 & 4\,880\,072 & 9\,830\,600  & 8\,603\,848 \\
\hline
Backbone &  \multicolumn{2}{c|}{22\,050\,664}&\multicolumn{2}{c|}{86\,567\,656}& \multicolumn{2}{c}{304\,368\,640} \\
\hline
\end{tabular}
%}
\label{tab:param_counts}
\end{table}
The linear probe and the attentive probe follow fundamentally different scaling behaviors. Given the hidden dimension $d$, the number of layers $|\mathcal{L}|$, and the number of classes $K$, the linear probe based on concatenation requires $ 2\cdot|\mathcal{L}|\cdot d\cdot K + K$ parameters, scaling linearly with both the number of layers and the number of classes. In contrast, ALF requires $8\cdot d^2+ 10d +d\cdot K + K$ parameters, which scales quadratically with the embedding dimension $d$, linearly with the number of classes, and remains independent of the number of layers used. Although the parameter count of the attention probe across all final-layer patches (AAT) is the same, its larger number of input tokens results in higher computational costs.

\Cref{tab:param_counts} compares parameter counts across three Vision Transformer architectures over a range of class counts representative of the datasets in our experiments. While the attentive probe has a higher fixed overhead, its class-dependent growth is substantially slower than that of concatenation. As the class count increases, the linear probe grows rapidly, whereas the attentive probe remains relatively stable. In practice, ALF uses fewer than $5\%$ of the backbone's parameters, offering a highly efficient solution that scales well to large multi-class problems.

\subsection{Importance of including structural information} \label{app:ap-token-inclusion}
We analyze the effect of token selection in ALF. We compare three configurations: attentive layer fusion using only \CLS tokens, encoding the semantic information, only \AP tokens, capturing more structural information by averaging spatial features, or both token types from all layers.

\begin{figure}
    \centering
    \includegraphics[width=\linewidth]{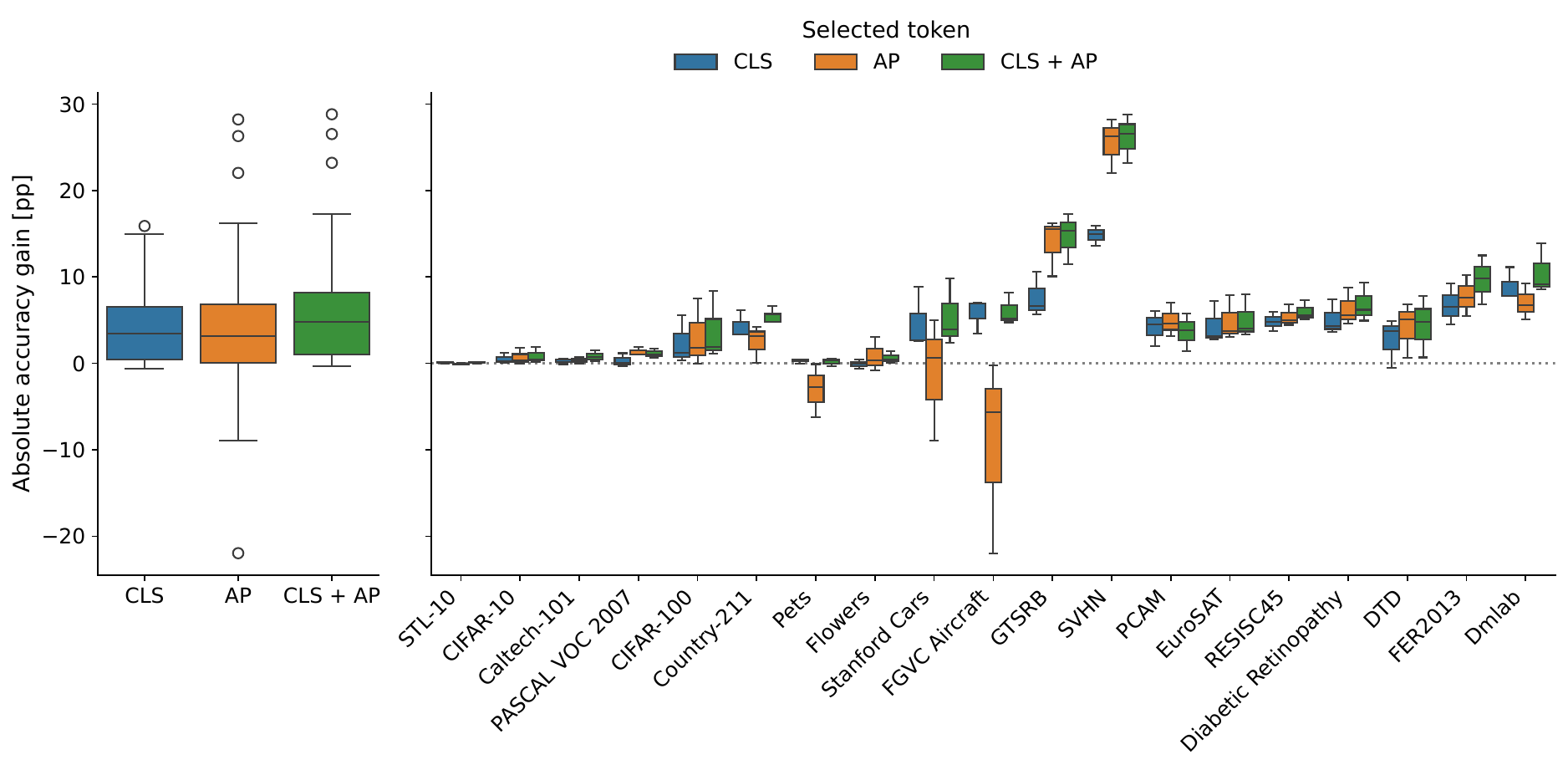}
    \caption{Absolute performance gains of attention-based intermediate layer fusion using different token configurations. Left: Distribution of gains across three base models and 19 datasets. Right: Per-dataset breakdown showing dataset-specific patterns in token utility.}
    \label{fig:ap-token-inclusion}
\end{figure}

\Cref{fig:ap-token-inclusion} shows absolute performance gains relative to the last layer \CLS linear probe baseline across the three base models (CLIP-B-16, DINOv2-B-14, ViT-B-16) on 19 datasets. We set the attention dropout to 0.1 to reduce the complexity of hyperparameter search.

The results demonstrate three key findings: (1) \CLS tokens consistently provide positive gains across most datasets, (2) \AP tokens exhibit high variance, substantially improving performance on some datasets (e.g., SVHN, GTSRB) while degrading it on others (e.g., FGVC Aircraft, Pets), and (3) combining both token types achieves the best overall performance, indicating the attention mechanism successfully learns when to utilize each token type.

\subsection{Risk of Overfitting}
\label{app:overfitting}
\begin{figure}[t]
    \centering
    \includegraphics[width=0.7\linewidth]{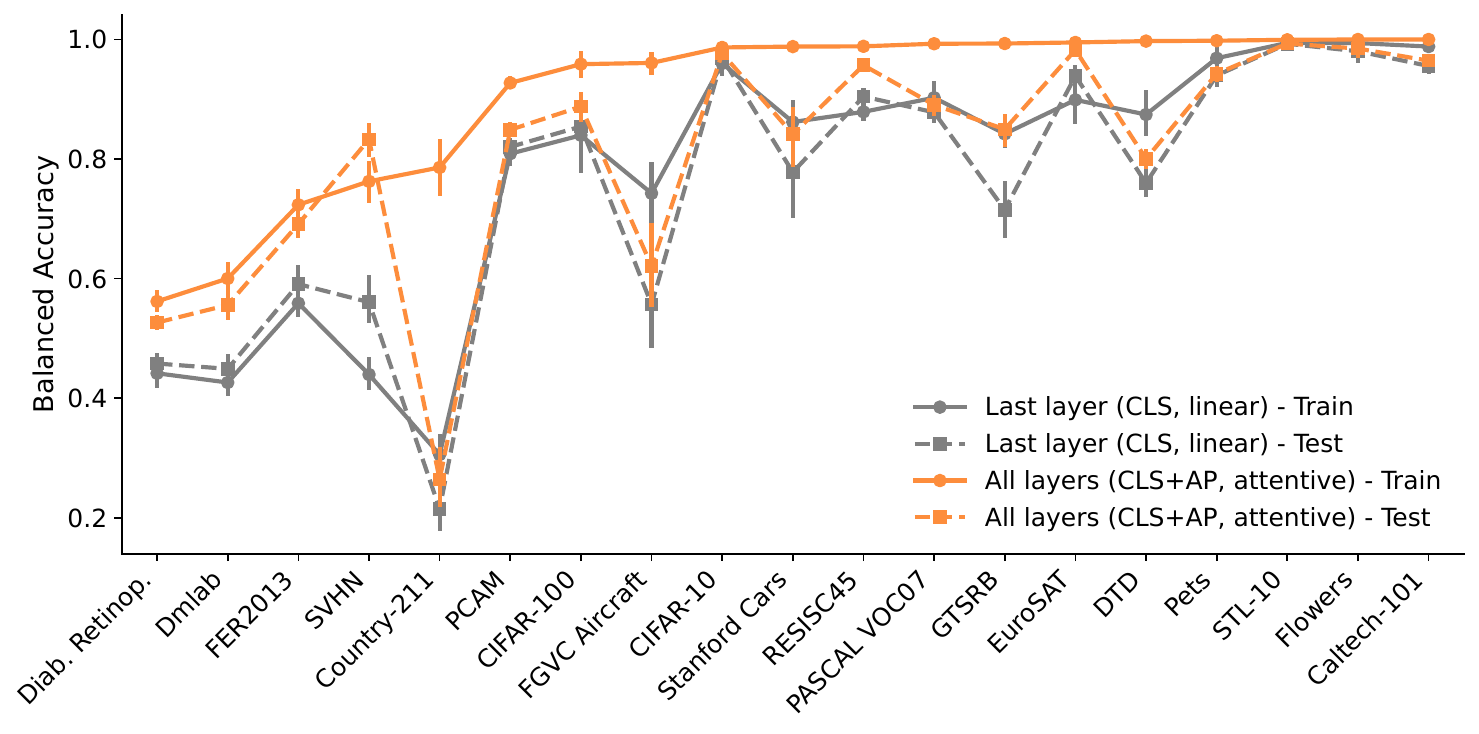}
    \caption{Train and test balanced accuracy comparison for each benchmark dataset across 9 models. The baseline performance (linear probe on last layer's \CLS token) versus attentive probe on \CLS+\AP of all intermediate layers are shown. While most datasets show acceptable overfitting patterns, PCAM and PASCAL VOC 2007 exhibit overfitting where the attentive method's test performance approaches the linear baseline despite higher training accuracy.}
    \label{fig:train-test-bal-acc-curves}
\end{figure}

More expressive probes inherently increase overfitting risk due to their greater capacity to memorize training-specific patterns. Despite mitigation strategies including weight decay and representational jittering, \Cref{fig:train-test-bal-acc-curves} reveals two overfitting patterns across our benchmark.

For most datasets, both methods exhibit similar train-test gaps, with our attentive fusion method maintaining superior test performance despite having a higher capacity. This represents acceptable overfitting where the additional expressiveness provides genuine benefits even with regularization. However, we observe overfitting on PCAM and PASCAL VOC 2007, where the linear baseline shows small train-test gaps while ALF overfits significantly despite regularization, resulting in test performance comparable to the simpler baseline (\Cref{tab:perf_gain_per_dataset}).

PCAM exemplifies this failure mode, potentially due to substantially more training updates (5{,}120 vs.\ 1{,}320 for our second-largest dataset) that may amplify overfitting effects. Additionally, standard data augmentation techniques could not be applied as regularization since we work with pre-extracted frozen features. Finally, the attentive fusion mechanism appears to overfit to noise in intermediate features, particularly from the \AP token, which dilutes localized signals through spatial averaging, problematic since PCAM's diagnostic information concentrates in small tissue regions. By contrast, AAT avoids this issue despite a similar parameter count, as its attention mechanism operates only on the final layer and can thus focus directly on central patches. By contrast, AAT avoids this issue despite a similar parameter count, as its attention mechanism operates only on the final layer.

This highlights a boundary condition: when label-relevant information is highly localized, \AP-based aggregation becomes suboptimal, and limiting training steps becomes crucial even with regularization.

\subsection{Identifying the optimal number of Heads} \label{app:nr-heads}
To determine the optimal number of attention heads for ALF, we conducted experiments using the DinoV2-B-16 model with all layers (CLS+AP, attentive pooling). While \cite{chen2024context} used 8 attention heads by default, we systematically evaluated different head configurations to identify the best setting for ALF.

Due to computational constraints, we performed this analysis on a subset of 8 datasets: Stanford Cars, Country-211, GTSRB, CIFAR-100, DTD, EuroSAT, Pets, and SVHN. The experimental setup differed slightly from our main experiments by removing attention dropout, jitter, and gradient clipping to isolate the effect of the number of heads.

\begin{figure}[ht]
\centering
\includegraphics[width=0.8\linewidth]{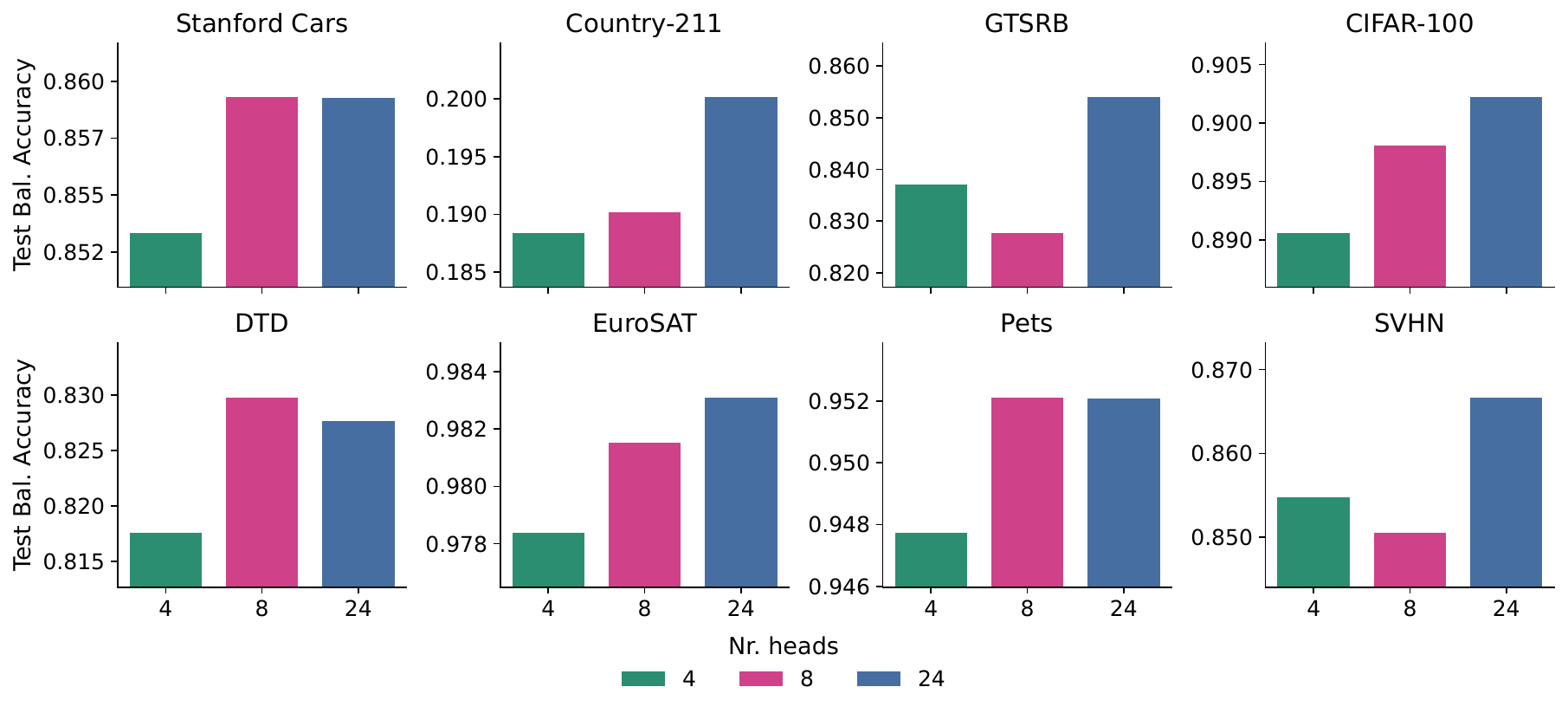}
\caption{Test balanced accuracy across different numbers of attention heads on 8 datasets, showing optimal performance when heads equal representations fused.}
\label{fig:nr_heads}
\end{figure}

\Cref{fig:nr_heads} shows that optimal performance is achieved when the number of attention heads equals the number of representations being fused, which we adopt for ALF.

\subsection{Comparison with State-of-the-Art Baselines}
\label{app:sota_comparison}

We evaluate ALF against two recent state-of-the-art adaptation methods that also rely on intermediate features: Head2Toe \citep{evci2022head2toe} and VQT \citep{tu2023vqt}. 

We adopted the evaluation setting reported in \citet{tu2023vqt} in order to compare with their reported results. Specifically, we retrained ALF using the same ViT-B-16 backbone (pre-trained strictly on ImageNet-1K, rather than our standard ImageNet-21K $\rightarrow$ 1K pipeline). Furthermore, we removed our class-balancing loss to report standard, unbalanced top-1 accuracy.

As shown in Table~\ref{tab:sota_baselines}, ALF significantly outperforms previous aggregation methods across the vast majority of datasets. On average, we achieve an \textbf{8.64[pp]} accuracy improvement over Head2Toe and \textbf{5.61[pp]} over VQT. Furthermore, unlike VQT, which requires continuous forward passes through the backbone during training, ALF utilizes fixed representations that can be pre-computed. This drastically reduces GPU memory consumption during training, demonstrating that our simple attentive probe is not only a more accurate mechanism for combining intermediate representations, but a more efficient one as well.

\begin{table*}[ht]
\centering
\caption{Comparison of adaptation methods using intermediate features. All models use the same ViT-B-16 (ImageNet-1K) backbone. Unbalanced top-1 accuracy is reported. ALF outperforms the baselines on 10 out of 11 datasets.}
\label{tab:sota_baselines}
\resizebox{\textwidth}{!}{%
\begin{tabular}{lccccccccccc}
\toprule
\textbf{Method} & \textbf{CIFAR-100} & \textbf{Caltech-101} & \textbf{DTD} & \textbf{Flowers} & \textbf{Pets} & \textbf{SVHN} & \textbf{PCAM} & \textbf{EuroSAT} & \textbf{RESISC45} & \textbf{Diab. Retionop.} & \textbf{Dmlab} \\ \midrule
%Fine-tuning* & 44.3 & 84.5 & 54.1 & 84.7 & 74.7 & 87.2 & 85.3 & 95.0 & 76.0 & 70.4 & 46.9 \\
Head2Toe* & 54.4 & 86.8 & 64.1 & 83.4 & 82.6 & 78.9 & 81.3 & 95.4 & 81.2 & 73.7 & 41.5 \\
VQT* & 58.4 & 89.4 & 66.7 & 90.4 & 89.1 &\textbf{ 81.1 }& 82.2 & 96.2 & 84.7 & 74.9 & 43.5 \\
\midrule
\textbf{ALF (Ours)} & \textbf{83.56} & \textbf{91.24} & \textbf{71.97} & \textbf{92.54} & \textbf{92.72} & 80.09 & \textbf{83.71} & \textbf{98.50} & \textbf{94.83} & \textbf{78.06} & \textbf{51.10} \\ \bottomrule
\multicolumn{12}{l}{\footnotesize *Results taken from \citet{tu2023vqt}.}
\end{tabular}%
}
\end{table*}

\subsection{Finetuning comparison}\label{app:finetuning}
Our work focuses on the probing paradigm, where the pretrained backbone remains frozen and only a lightweight classification head is trained. This approach is valuable in resource-constrained scenarios or when the model must serve multiple tasks and should therefore not be changed.
However, to contextualize our contributions within the broader landscape of transfer learning methods, we conducted additional fine-tuning and LoRA experiments \citep{hu2022lora} for the three base models (CLIP-B-16, DINOv2-B-14, and ViT-B-16) on GTSRB, CIFAR-100, and EuroSAT.
Due to computational constraints, we used fixed hyperparameters for each model and dataset: learning rate $1 \times 10^{-3}$ and weight decay $1 \times 10^{-1}$ for the classification head, learning rate $1 \times 10^{-5}$ and weight decay $1 \times 10^{-6}$ for the backbone. We train for 40 epochs with a batch size of 256, enforcing at least 1000 gradient updates as in our main experiments.

\begin{figure}[ht]
\centering
\includegraphics[width=1\linewidth]{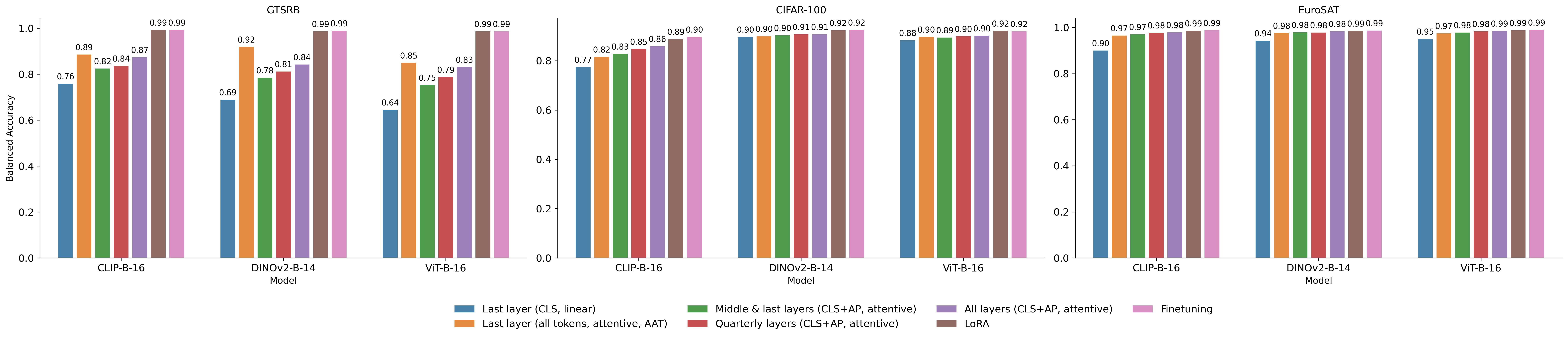}
\caption{Downstream performance of probing strategies, finetuning, and LoRA for three datasets (GTSRB, CIFAR-100, and EuroSAT) and the three base models.}
\label{fig:fine_tuning}
\end{figure}

\begin{figure}[ht]
\centering
\includegraphics[width=\linewidth]{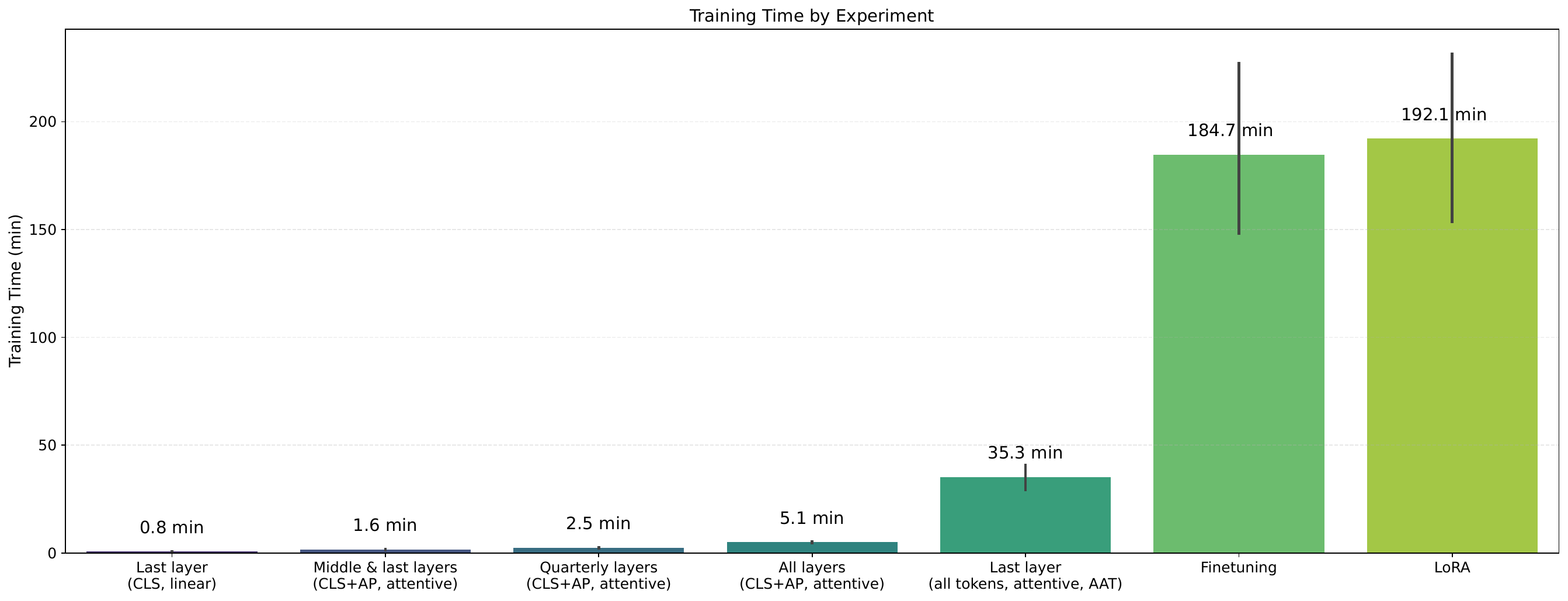}
\caption{Training times in minutes for probing strategies and finetuning averaged across datasets and the three base models.}
\label{fig:training_times}
\end{figure}

\Cref{fig:fine_tuning} reveals that while fine-tuning and LoRA generally achieve the highest accuracy, ALF closely matches its performance on CIFAR-100 and EuroSAT. On GTSRB, fine-tuning achieves substantially higher accuracy (7-15pp), reflecting the value of direct backbone adaptation for fine-grained spatial discrimination. Importantly, ALF consistently outperforms linear probing across all datasets while being 36 times faster (5.1 min vs 184.7 min) during training compared to fine-tuning (\Cref{fig:training_times}), demonstrating a practical accuracy-efficiency trade-off for resource-constrained scenarios where probing is preferred.

\subsection{Multi-layer fusion across attention probe architectures}\label{app:probe_comparison}
\begin{figure}
    \centering
    \includegraphics[width=\linewidth]{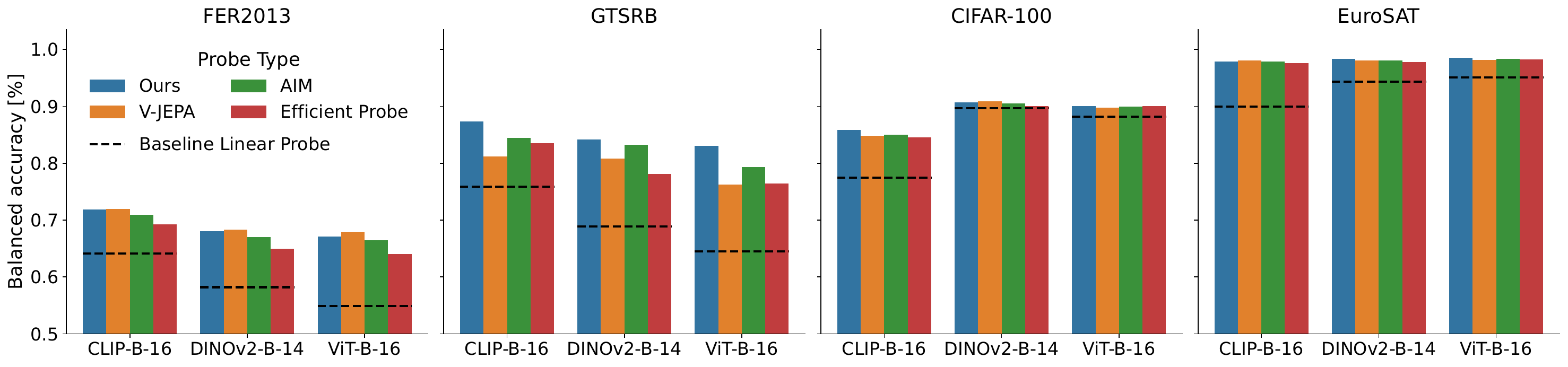}
    \caption{Performance of multi-layer attentive fusion using different per-layer attention probes. We compare our CAE-style probe with V-JEPA, AIM, and Efficient Probe (EP), using both \CLS and \AP tokens from all layers.
    }
    \label{fig:probe_comparison}
\end{figure}
To verify that the benefits of multi-layer attentive fusion do not depend on the specific design of the attention module, we evaluate several alternative attentive probes in place of our CAE-style implementation~\cite{chen2024context}. Specifically, we consider AIM~\cite{el2024scalable}, Efficient Probe (EP)~\cite{psomas2025attentionpleaserevisitingattentive}, and V-JEPA~\cite{bardes2024revisiting}, assessing their ability to aggregate intermediate layer information.

\Cref{fig:probe_comparison} reports accuracy on four representative datasets. All attentive probe variants consistently outperform the standard last-layer \CLS linear probe, confirming that multi-layer attentive fusion effectively leverages intermediate representations independent of the attention design. Differences between probe types are minor, with more complex probes (CAE, V-JEPA) showing slightly higher gains on some tasks than the simpler variants (EP, AIM). In summary, the results confirm that multi-layer attentive fusion provides consistent downstream benefits across probe architectures, reinforcing the generality of ALF and validating the key claim that intermediate-layer features contain task-relevant information beyond the final layer \CLS and \AP tokens.

\subsection{Stability of experiment runs}\label{app:stability-seeds}
\begin{figure}[t]
    \centering
    \includegraphics[width=0.85\linewidth]{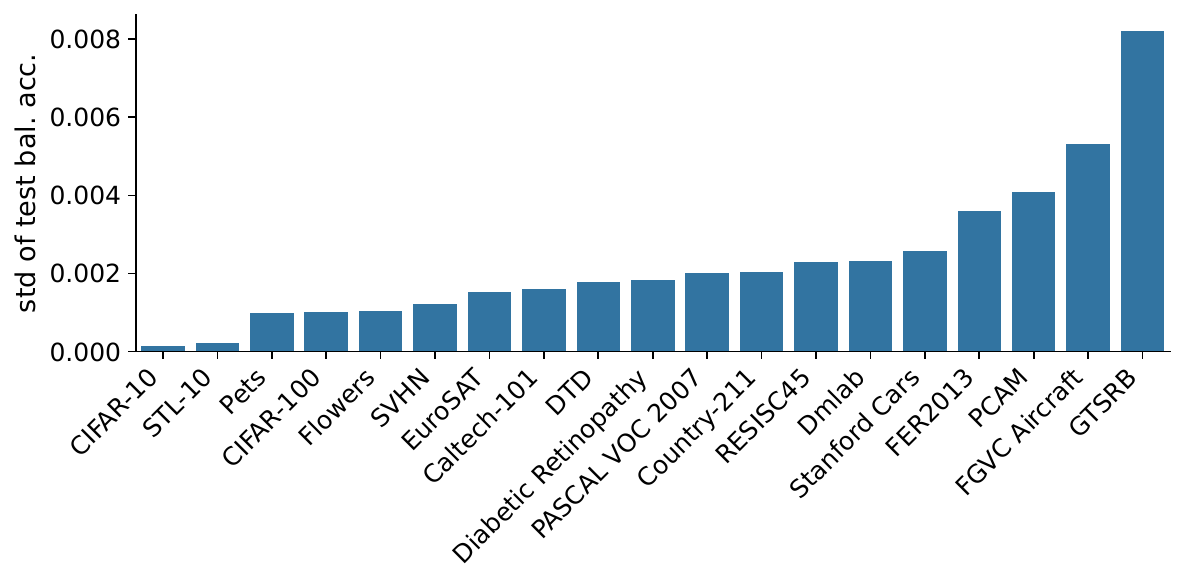}
    \caption{Standard deviation of balanced test accuracy across five random seeds for DinoV2-B-14 with all layers (CLS+AP, attentive pooling) on 19 datasets. All values remain below 0.01, indicating stable performance across different random initializations.}
    \label{fig:app-var-seeds}
\end{figure}

To assess the stability of our experimental results, we conducted a seed variation analysis using the DinoV2-B-16 model with all layers (CLS+AP, attentive pooling). We ran five different random seeds for 19 datasets used in our evaluation. To reduce the hyperparameter search space, we removed attention dropout and focused the tuning process on learning rate and weight decay only.
\Cref{fig:app-var-seeds} shows the standard deviation of balanced test accuracy across the five seeds for each dataset. The results demonstrate that standard deviation remains below 0.01 for all datasets, with many datasets achieving standard deviations below 0.002. These values indicate that the variance across different random seeds is limited.
Based on this stability analysis, we determined that single runs for each dataset and configuration would be sufficient for our main experiments, enabling us to allocate computational resources more efficiently while maintaining reliable results.

\subsection{Use of Large Language Models}
Large language models (Google's Gemini, OpenAI's ChatGPT, and Anthropic's Claude) were used as a writing assistant to help refine the language and improve the clarity of the manuscript. Separately, AI-powered coding tools like Cursor and GitHub Copilot were used for advanced auto-completion during software development. The human authors directed all scientific aspects of the work, including the research ideas, methodology, and analysis of results, and are fully responsible for the content of the paper.

\end{document}